\documentclass{article}

\usepackage{microtype}
\usepackage{booktabs} 
\usepackage{url}
\usepackage[pdftex]{graphicx}
\usepackage{amsthm}
\usepackage{amsmath}
\usepackage{amssymb}
\usepackage{mathtools}
\usepackage{comment}
\usepackage{algorithm,algorithmic}
\usepackage{float}
\usepackage{times}
\usepackage{caption}
\usepackage{chngcntr}
\usepackage{multirow}
\usepackage[skip=0cm,list=true,labelfont=it]{subcaption}

\DeclarePairedDelimiterX{\infdivx}[2]{(}{)}{%
  #1\;\delimsize\|\;#2%
}

\newtheorem{definition}{Definition}
\newtheorem{example}{Example}
\newcommand{\mX}{\mathbf{X}}
\newcommand{\mY}{\mathbf{Y}}
\newcommand{\mI}{\mathbf{I}}
\newcommand{\R}{\mathbb{R}}

\usepackage[breaklinks=true]{hyperref}

\usepackage[accepted]{icml2020}

\icmltitlerunning{Certified Robustness to Label-Flipping Attacks via Randomized Smoothing}

\begin{document}

\twocolumn[
\icmltitle{Certified Robustness to Label-Flipping Attacks via Randomized Smoothing}



\icmlsetsymbol{equal}{*}

\begin{icmlauthorlist}
\icmlauthor{Elan Rosenfeld}{cmu}
\icmlauthor{Ezra Winston}{cmu}
\icmlauthor{Pradeep Ravikumar}{cmu}
\icmlauthor{J. Zico Kolter}{cmu,bosch}
\end{icmlauthorlist}

\icmlaffiliation{cmu}{Carnegie Mellon University}
\icmlaffiliation{bosch}{Bosch Center for AI}

\icmlcorrespondingauthor{Elan Rosenfeld}{elan@cmu.edu}

\icmlkeywords{Machine Learning, ICML}

\vskip 0.3in
]



\printAffiliationsAndNotice{}  

\begin{abstract}
Machine learning algorithms are known to be susceptible to data poisoning attacks, where an adversary manipulates the training data to degrade performance of the resulting classifier. In this work, we present a unifying view of randomized smoothing over arbitrary functions, and we leverage this novel characterization to propose a new strategy for building classifiers that are \emph{pointwise-certifiably robust} to general data poisoning attacks. As a specific instantiation, we utilize our framework to build linear classifiers that are robust to a strong variant of label flipping, where each test example is targeted independently. In other words, for each test point, our classifier includes a certification that its prediction would be the same had some number of training labels been changed adversarially. Randomized smoothing has previously been used to guarantee---with high probability---test-time robustness to adversarial manipulation of the input to a classifier; we derive a variant which provides a deterministic, analytical bound, sidestepping the probabilistic certificates that traditionally result from the sampling subprocedure. Further, we obtain these certified bounds with minimal additional runtime complexity over standard classification and no assumptions on the train or test distributions. We generalize our results to the multi-class case, providing the first multi-class classification algorithm that is certifiably robust to label-flipping attacks.
\end{abstract}

\section{Introduction}

Modern classifiers, despite their widespread empirical success, are known to be susceptible to adversarial attacks. In this paper, we are specifically concerned with so-called ``data poisoning'' attacks (formally, \emph{causative} attacks [\citealt{MLSecure,PapernotSoK}]), where the attacker manipulates some aspects of the training data in order to cause the learning algorithm to output a faulty classifier. Automated machine-learning systems which rely on large, user-generated datasets---e.g. email spam filters, product recommendation engines, and fake review detectors---are particularly susceptible to such attacks. For example, by maliciously flagging legitimate emails as spam and mislabeling spam as innocuous, an adversary can trick a spam filter into mistakenly letting through a particular email.

Data poisoning attacks in the literature include label-flipping attacks \citep{LabelFlipSVM}, where the labels of a training set can be adversarially manipulated to decrease performance of the trained classifier; general data poisoning, where both the training inputs and labels can be manipulated \citep{SteinhardtCertified}; and backdoor attacks \citep{Backdoor, SpectralBackdoor}, where the training set is corrupted so as to cause the classifier to deviate from its expected behavior only when triggered by a specific pattern. However, unlike the alternative test-time adversarial setting, where reasonably effective provable defenses exist, comparatively little work has been done on building classifiers that are certifiably robust to targeted data poisoning attacks.

In this work we propose a framework for building classifiers that are certifiably robust to a given class of data poisoning attacks, such as label-flipping or backdoor attacks. In particular, we propose what we refer to as a \emph{pointwise} certified defense---this means that with each prediction, the classifier includes a certificate guaranteeing that its prediction would not be different had it been trained on adversarially manipulated data up to some ``radius of perturbation" (a formal definition is presented in Section~\ref{sec:general-smoothing}). We then demonstrate a specific instantiation of this protocol, constructing linear classifiers that are pointwise-certifiably robust to label-flipping attacks; i.e., each prediction is certified robust against a certain number of training label flips.

Prior works on certified defenses make statistical guarantees over the entire test distribution, but they make no guarantees as to the robustness of a prediction on any particular test point; thus, a determined adversary could still cause a specific test point to be misclassified. We therefore consider the threat of a worst-case adversary that can make a training set perturbation to target \emph{each test point individually}. This motivates a defense that can certify each of its individual predictions, as we present here. Compared to traditional robust classification, this framework is superior for a task such as determining who receives a coveted resource (a loan, parole, etc.), as it provides a guarantee for each individual, rather than at the population level. This work represents the first such pointwise certified defense to \emph{any type of data poisoning attack}; we expect significant advances can be made on both attacks and defenses within this threat model.

Our approach leverages randomized smoothing \citep{Cohen}, a technique that has previously been used to guarantee test-time robustness to adversarial manipulation of the input to a deep network. However, where prior uses of randomized smoothing randomize over the input to the classifier for test-time guarantees, we instead randomize over \emph{the entire training procedure of the classifier}. Specifically, by randomizing over the labels during this training process, we obtain an overall classification pipeline that is certified to not change its prediction when some number of labels are adversarially manipulated in the training set. Previous applications of randomized smoothing perform sampling to provide probabilistic bounds, due to the intractability of integrating the decision regions of a deep network. We instead derive an analytical bound, providing truly guaranteed robustness. Although a naive implementation of this approach would not be tractable, we show how to obtain these certified bounds with minimal additional runtime complexity over standard classification, suffering only a linear cost in the number of training points.

A further distinction of our approach is that the applicability of our robustness guarantees do not rely upon stringent model assumptions or the quality of the features. Existing work on robust linear classification or regression provides certificates that only hold under specific model assumptions, e.g., recovering the best-fit linear coefficients, which is most useful when the data exhibit a linear relationship in the feature space. In contrast, our classifier makes \textit{no assumptions} about the separability of the data or quality of the features; this means our certificates remain valid when applying our classifier to arbitrary features, which in practice allows us to leverage advances in unsupervised feature learning \citep{UnsupervisedFeatures, SimCLR} and transfer learning \citep{Donahue}. We apply our classifier to pre-trained and unsupervised deep features to demonstrate its feasibility for classification of highly non-linear data such as ImageNet.

We evaluate our proposed classifier on several benchmark datasets common to the data poisoning literature. On the Dogfish binary classification challenge from ImageNet, our classifier maintains 81.3\% certified accuracy in the face of an adversary who could reduce an undefended classifier to less than 1\%. Additional experiments on MNIST and CIFAR10 demonstrate our algorithm's effectiveness for multi-class classification. Moreover, our classifier maintains a reasonably competitive non-robust accuracy (e.g., 94.5\% on MNIST 1/7 versus 99.1\% for the undefended classifier).

\section{Related Work}

\paragraph{Data poisoning attacks}
A \emph{data poisoning attack} \citep{BackGradientPoisoning, PoisonNeural} is an attack where an adversary corrupts some portion of the training set or adds new inputs, with the goal of degrading the performance of the learned model. The adversary is assumed to have perfect knowledge of the learning algorithm, so security by \emph{design}---as opposed to obscurity---is the only viable defense against such attacks. The adversary is also typically assumed to have access to the training set and, in some cases, the test set.

Previous work has investigated attacks and defenses for data poisoning attacks applied to feature selection \citep{XiaoFeatureSelection}, SVMs \citep{BiggioSVM, LabelFlipSVM}, linear regression \citep{RobustLinearRegPoison}, and PCA \citep{PCAPoison}, to name a few. Some attacks can even achieve success with ``clean-label" attacks, inserting adversarially perturbed, seemingly correctly labeled training examples that cause the classifier to perform poorly \citep{PoisonFrogs, CleanLabel}. Interestingly, our defense can also be viewed as (the first) certified defense to such attacks: perturbing an image such that the resulting features no longer match the label is theoretically equivalent to changing the label such that it no longer matches the image's features. For an overview of data poisoning attacks and defenses in machine learning, see \citet{BiggioPoisonSurvey}.

\paragraph{Label-flipping attacks}
A \emph{label-flipping attack} is a specific type of data poisoning attack where the adversary is restricted to changing the training labels. The classifier is then trained on the corrupted training set, with no knowledge of which labels have been tampered with. For example, an adversary could mislabel spam emails as innocuous, or flag real product reviews as fake.

Unlike random label noise, for which many robust learning algorithms have been successfully developed \citep{NoisyLabels, LiuReweighting, NNLabelNoise}, adversarial label-flipping attacks can be specifically targeted to exploit the structure of the learning algorithm, significantly degrading performance. Robustness to such attacks is therefore harder to achieve, both theoretically and empirically \citep{LabelFlipSVM, BiggioSVM}. A common defense technique is \emph{sanitization}, whereby a defender attempts to identify and remove or relabel training points that may have had their labels corrupted \citep{Sanitization, RahimMalware}. Unfortunately, recent work has demonstrated that this is often not enough against a sufficiently powerful adversary \citep{KohBreakSanitization}. Further, \emph{no existing defenses} provide pointwise guarantees regarding their robustness.

\paragraph{Certified defenses} Existing works on certified defenses to adversarial data poisoning attacks typically focus on the regression case and provide broad statistical guarantees over the entire test distribution. A common approach to such certifications is to show that a particular algorithm recovers some close approximation to the best linear fit coefficients \citep{RobustLinearRegression, RobustGradient, TrimmedLoss}, or that the expected loss on the test distribution is bounded \citep{EfficientAlgorithms, DRO}. These results generally rely on assumptions on the data distribution: some assume sparsity in the coefficients \citep{CompressedSensing, ICMLRobustSparseRegression} or corruption vector \citep{HardThresholding}; others require limited effects of outliers \citep{SteinhardtCertified}. As mentioned above, all of these methods fail to provide guarantees for individual test points. Additionally, these statistical guarantees are not as meaningful when their model assumptions do not hold.

\paragraph{Randomized smoothing} Since the discovery of adversarial examples \citep{SzegedyIntriguing, GoodfellowAdversarial}, the research community has been investigating techniques for increasing the adversarial robustness of complex models such as deep networks. After a series of heuristic defenses, followed by attacks breaking them \citep{obfuscated-gradients, CarliniDetection}, focus began to shift towards the development of \emph{provable} robustness.

One approach which has gained popularity in recent work is randomized smoothing. Rather than certifying the original classifier $f$, randomized smoothing defines a new classifier $g$ whose prediction at an input $x$ is the class assigned the most probability when $x$ is perturbed with noise from some distribution $\mu$ and passed through $f$. That is, $g(x) = \arg\max_c \; \mathbb{P}_{\epsilon\sim\mu}(f(x+\epsilon) = c).$ This new classifier $g$ is then certified as robust, ideally without sacrificing too much accuracy compared to $f$. The original formulation was presented by \citet{Lecuyer} and borrowed ideas from differential privacy. The above definition is due to \citet{Li} and was popularized by \citet{Cohen}, who derived a tight robustness guarantee.

\section{A General View of Randomized Smoothing}
\label{sec:general-smoothing}
Our first contribution is a general viewpoint of randomized smoothing, unifying all existing applications of the framework. Under our notation, randomized smoothing constructs an operator $G(\mu, \phi)$ that maps a binary-valued\footnote{For simplicity, we present the methodology here with binary-valued functions, which will correspond eventually to binary classification problems. The extension to the multi-class setting requires additional notation, and thus is deferred to the appendix.} function $\phi : \mathcal{X} \to \{0,1\}$ and a \emph{smoothing measure} $\mu : \mathcal{X} \to \mathbb{R_+}$, with $\int_\mathcal{X}\mu(x)dx=1$, to the expected value of $\phi$ under $\mu$ (that is, $G(\mu, \phi)$ represents the ``vote" of $\phi$ weighted by $\mu$). For example, $\phi$ could be a binary image classifier and $\mu$ could be some small, random pixel noise applied to the to-be-classified image. We also define a ``hard threshold" version $g(\mu, \phi)$ that returns the most probable output (the majority vote winner). Formally,
\begin{align*}
    G(\mu, \phi) &= \mathbb{E}_{x \sim \mu}[\phi(x)] = \int_\mathcal{X} \mu(x) \phi(x) dx, \\
    g(\mu, \phi) &= \mathbf{1}\{G(\mu, \phi) \geq 1/2\},
\end{align*}

where $\mathbf{1}\{\cdot\}$ is the indicator function. Where it is clear from context, we will omit the arguments, writing simply $G$ or $g$. Intuitively, for two similar measures $\mu,\rho$, we would expect that for most $\phi$, even though $G(\mu, \phi)$ and $G(\rho, \phi)$ may not be equal, the threshold function $g$ should satisfy $g(\mu, \phi) = g(\rho, \phi)$.  Further, the degree to which $\mu$ and $\rho$ can differ while still preserving this property should increase as $G(\mu, \phi)$ approaches either 0 or 1, because this increases the ``margin'' with which the function $\phi$ is 0 or 1 respectively over the measure $\mu$.  More formally, we define a general randomized smoothing guarantee as follows:

\begin{definition}
Let $\mu : \mathcal{X} \to \mathbb{R_+}$ be a smoothing measure over $\mathcal{X}$, with $\int_{\mathcal{X}}\mu(x)dx = 1$.\footnote{There is no theoretical reason to restrict $\mu$ to be a probability measure. While this and all previous works only consider probability measures, the framework we present here could easily be extended to allow for more general measures $\mu, \rho$ and functions $\phi$.} Then a randomized smoothing robustness guarantee is a specification of a distance function over probability measures $d(\mu, \rho)$ and a function $f : [0, 1] \to \mathbb{R}_+$ such that for all $\phi : \mathcal{X} \to \{0,1\}$,
\begin{equation}
\label{eq:robustness-guarantee}
    g(\mu, \phi) = g(\rho, \phi) \;\; \mbox{whenever} \;\; d(\mu, \rho) \leq f(G(\mu, \phi)).
\end{equation}
\end{definition}

Informally, \eqref{eq:robustness-guarantee} says that the majority vote winner of $\phi$ weighted by $\mu$ and $\rho$ will be the same, so long as $\mu$ and $\rho$ are ``close enough" as a function of the margin with which the majority wins. We will sometimes use $p$ in place of $G(\mu, \phi)$, representing the fraction of the vote that the majority class receives (analogous to $p_A$ in \citet{Cohen}).

\paragraph{Instantiations of randomized smoothing} This definition is rather abstract, so we highlight concrete examples of how it can be applied to achieve certified guarantees against adversarial attacks.

\begin{example}
The randomized smoothing guarantee of \citet{Cohen} uses the smoothing measures $\mu = \mathcal{N}(x_0,\sigma^2 I)$, a Gaussian aroound the point $x_0$ to be classified, and $\rho = \mathcal{N}(x_0+\delta,\sigma^2 I)$, the same measure perturbed by $\delta$. They prove that (\ref{eq:robustness-guarantee}) holds for all classifiers $\phi$ if we define
\begin{equation*}
    d(\mu, \rho) = \frac{1}{\sigma} \|\delta\|_2 \equiv \sqrt{2 \mathrm{KL}\infdivx{\mu}{\rho}}, \;\; f(p) = |\Phi^{-1}(p)|,
\end{equation*}
where $\mathrm{KL}(\cdot)$ denotes KL divergence and $\Phi^{-1}$ denotes the inverse CDF of the Gaussian distribution.
\end{example}

Although this work focused on the case of randomized smoothing of continuous data via Gaussian noise, this is by no means a requirement. \citet{Stratified} consider an alternative approach for dealing with discrete variables.
\begin{example}
The randomized smoothing guarantee of \citet{Stratified} uses the factorized smoothing measure in $d$ dimensions $\mu_{\alpha,K}(\mathbf{x}) = \Pi_{i=1}^d\mu_{\alpha,K,i}(\mathbf{x}_i)$, defined with respect to parameters $\alpha\in[0,1],K\in\mathbb{N}$, and a base input $\mathbf{z} \in \{0,\ldots,K\}^d$, where
\begin{align*}
    \mu_{\alpha,K,i}(\mathbf{x}_i) =
    \begin{cases}
    \alpha, &\text{if $\mathbf{x}_i=\mathbf{z}_i$}\\
    \frac{1-\alpha}{K}, &\text{if $\mathbf{x}_i\in\{0,\ldots,K\},\ \mathbf{x}_i\neq \mathbf{z}_i$},
    \end{cases}
\end{align*}
with $\mathbf{x}_i$ being the $i^{th}$ dimension of $\mathbf{x}$. $\rho_{\alpha,K}$ is similarly defined for a perturbed input $\mathbf{z}'$. They guarantee that (\ref{eq:robustness-guarantee}) holds if we define
\begin{equation}
\label{eq:stratified-bound}
    d(\mu,\rho) = \|\mathbf{z}'-\mathbf{z}\|_0, \;\; f(p) = \mathcal{F}_{\alpha,K,d}(\max(p, 1-p)).
\end{equation}
\end{example}
In words, the smoothing distribution is such that each dimension is independently perturbed to one of the other $K$ values uniformly at random with probability $1- \alpha$. $\mathcal{F}_{\alpha,K,d}(p)$ is a combinatorial function defined as the maximum number of dimensions---out of $d$ total---by which $\mu_{\alpha,K}$ and $\rho_{\alpha,K}$ can differ such that a set with measure $p$ under $\mu_{\alpha,K}$ is guaranteed to have measure at least $\frac{1}{2}$ under $\rho_{\alpha,K}$. \citet{Stratified} prove that this value depends only on $\|\mathbf{z}'-\mathbf{z}\|_0$.

Finally, \citet{DJ} consider a more general form of randomized smoothing that doesn't require strict assumptions on the distributions but is still able to provide similar guarantees.
\begin{example}[Generic bound]
    Given any two smoothing distributions $\mu, \rho$, we have the generic randomized smoothing robustness certificate, ensuring that (\ref{eq:robustness-guarantee}) holds with definitions
    \begin{equation}
        \label{eq:generic-bound}
        d(\mu,\rho) = \mathrm{KL}\infdivx{\rho}{\mu}, \;\; f(p) = -\frac{1}{2}\log(4p(1-p)).
    \end{equation}
\end{example}

\paragraph{Randomized smoothing in practice}
For deep classifiers, the expectation $G$ cannot be computed exactly, so we must resort to Monte Carlo approximation. This is done by drawing samples from $\mu$ and using these to construct a high-probability bound on $G$ for certification. More precisely, this bound should be a \emph{lower} bound on $G$ when the hard prediction $g = 1$ and an \emph{upper} bound otherwise; this ensures in both cases that we under-certify the true robustness of the classifier $g$.  The procedure is shown in Algorithm \ref{alg:generic-randomized-smoothing} in Appendix \ref{appendix:generic-randomized-smoothing}. These estimates can then be plugged into a randomized smoothing robustness guarantee to provide a high probability certified robustness bound for the classifier.

\section{Pointwise Data Poisoning Robustness}
We now present the main contributions of this paper: we first describe a generic strategy for applying randomized smoothing to certify a prediction function against arbitrary classes of data poisoning attacks. We then propose a specific implementation of said strategy to certify a classifier against label-flipping attacks. We show how this approach can be made tractable using linear least-squares classification, and we use the Chernoff inequality to analytically bound the relevant probabilities for the randomized smoothing certificate. Notably, although we are employing a randomized approach, the final algorithm does not use any random sampling, but rather relies upon a convex optimization problem to compute the certified robustness.

\label{sec:kl-bound}
\paragraph{General data poisoning robustness} We begin by noting that in prior work, randomized smoothing was applied at test time with the function $\phi : \mathcal{X} \to \{0,1\}$ being a (potentially deep) classifier that we wish to smooth.  However, there is no requirement that the function $\phi$ be a classifier at all; the theory holds for any binary-valued function. Instead of treating $\phi$ as a trained classifier, we consider $\phi$ to be \emph{an arbitrary learning algorithm} which takes as input a training dataset $\{x_i,y_i\}_{i=1}^n \in (\mathcal{X} \times \{0,1\})^n$ and additional test points without corresponding labels, which we aim to predict.\footnote{Note that our algorithm does not require access to the test data to do the necessary precomputation. We present it here as such merely to give an intuitive idea of the procedure.} In other words, the combined goal of $\phi$ is to first train a classifier and then predict the label of the new example. Thus, we consider test time outputs to be a function of both the test time input and the training data that produced the classifier. This perspective allows us to reason about how changes to training data affect the classifier at test time, reminiscent of work on influence functions of deep neural networks \citep{KohInfluence, RepresenterPoint}.

This immediately suggests our protocol for pointwise robustness to general data poisoning attacks: randomize over the elements of the input to which we desire certified robustness, rather than over the test-time input to be classified. For example, to induce robustness to backdoor attacks, we could randomly add noise to the training points and/or their labels. Analogous to previous applications of randomized smoothing, if the majority vote of the classifiers trained with these randomly perturbed inputs has a large margin, it will confer a degree of robustness within an appropriately-defined radius of adversarially perturbed training data (as defined in Equation~\ref{eq:robustness-guarantee}).

\paragraph{Specific application to label-flipping robustness}
To demonstrate the effectiveness of our proposed strategy, we now present a specific implementation, providing an algorithm for tractable linear classification which is pointwise-certifiably robust to label-flipping attacks. When applying randomized smoothing in this setting, we randomize over the labels in the training set as described above---a suitably large margin in the majority vote will therefore result in pointwise robustness to adversarial label flips. In this scenario, the adversarial ``radius" is defined as number of labels on which two training sets differ.

To formalize this intuition, consider two different assignments of $n$ training labels  $Y_1,Y_2 \in \{0,1\}^n$ which differ on precisely $r$ labels.  Let $\mu$ (resp. $\rho$) be the distribution resulting from independently flipping each of the labels in $Y_1$ (resp. $Y_2$) with probability $q$. It is clear that as $r$ increases, $\mathrm{KL}\infdivx{\mu}{\rho}$ will also increase. In fact, it is simple to show (see Appendix \ref{appendix:kl-bound-derivation} for derivation) that the exact KL divergence between these two distributions is
\begin{equation}
\label{eq:kl-divergence}
    \mathrm{KL}\infdivx{\mu}{\rho} = \mathrm{KL}\infdivx{\rho}{\mu} = r (1-2q)\log\left(\frac{1-q}{q}\right).
\end{equation}
Plugging in the robustness guarantee (\ref{eq:generic-bound}), we have that $g(\mu, \phi) = g(\rho, \phi)$ so long as
\begin{equation}
\label{eq:kl-bound}
    r \le \frac{\log(4p(1-p))}{2(1-2q)\log\left(\frac{q}{1-q}\right)},
\end{equation}
where $p = G(\mu, \phi)$.
This implies that for any test point, as long as (\ref{eq:kl-bound}) is satisfied, $g$'s prediction (the majority vote weighted by the smoothing distribution) will not change if an adversary corrupts the training set from $Y_1$ to $Y_2$, or indeed to any other training set that differs on at most $r$ labels. We can tune the noise hyperparameter $q$ to achieve the largest possible upper bound in (\ref{eq:kl-bound}); more noise will likely decrease the margin of the majority vote $p$, but it will also decrease the divergence.

\paragraph{Computing a tight bound} This approach has a simple closed form, but the bound is not tight. We can derive a tight bound via a combinatorial approach as in \citet{Stratified}. By precomputing the quantities $\mathcal{F}^{-1}_{1-q,1,n}(r)$ from Equation (\ref{eq:stratified-bound}) for each $r$, we can simply compare $p$ to each of these and thereby certify robustness to the highest possible number of label flips. This precomputation can be expensive, but it provides a significantly tighter robustness guarantee, certifying approximately twice as many label flips for a given bound on $G$ (See Figure \ref{fig:duality-gap} in the Appendix).

\subsection{Efficient implementation via least squares classifiers} \label{sect:efficient-implementation}

There may appear to be one major impracticality of the algorithm proposed in the previous section, if considered naively: treating the function $\phi$ as an entire training-plus-single-prediction process would require that we train multiple classifiers, over multiple random draws of the labels $y$, all to make a prediction on a single example. In this section, we describe a sequence of tools we employ to restrict the architecture and training process in a manner that drastically reduces this cost, bringing it in line with the traditional cost of standard classification. The full procedure, with all the parts described below, can be found in Algorithm \ref{alg:certified-label-flipping}.

\begin{algorithm}[t]
\caption{Randomized smoothing for label-flipping robustness}
\label{alg:certified-label-flipping}
\begin{algorithmic}
\STATE \textbf{Input:} feature mapping $h : \R^d \to \R^k$; noise parameter $q$; regularization parameter $\lambda$; training set $\{(x_i, y_i) \in \R^{d}\times\{0,1\}\}_{i=1}^n$ (with potentially adversarial labels); additional inputs to predict $\{x_j\in\R^d\}_{j=1}^m$.
\STATE 1. Pre-compute matrix $\mathbf{M}$,
\begin{equation*}
    \mathbf{M} = \mX \left (\mX^T \mX + \lambda \mI\right)^{-1}
\end{equation*}
where $\mX \equiv h(x_{1:n})$.
\FOR{$j = 1,\ldots,m$}
\STATE 1. Compute vector $\boldsymbol \alpha^j = \mathbf{M} h(x_{j})^T$.
\STATE 2. Compute optimal Chernoff parameter $t$ via Newton's method
\begin{multline*}
t^\star = \arg\min_t \biggl \{ t/2 + \sum_{i:y_i=1} \log\left(q + (1-q)e^{-t\boldsymbol \alpha^j_i}\right) \\ + \sum_{i:y_i=0} \log\left((1-q) + q e^{-t\boldsymbol \alpha^j_i}\right) \biggr \}
\end{multline*}
and let $p^\star = \max(1-B_{|t^\star|}, 1/2)$ where $B_{|t^\star|}$ is the Chernoff bound (\ref{eq:chernoff-bound}) evaluated at $|t^\star|$.
\STATE \textbf{Output:} Prediction $\hat{y}_{j} = \mathbf{1}\left\{t^\star\geq0\right\}$ and certification that prediction will remain constant for up to $r$ training label flips, where
\begin{equation*}
r = \left \lfloor{\frac{\log(4p^\star(1-p^\star))}{2(1-2q)\log\left(\frac{q}{1-q}\right)}}\right \rfloor.
\end{equation*}
\ENDFOR
\end{algorithmic}
\end{algorithm}

\paragraph{Linear least-squares classification} The fundamental simplification we make in this work is to restrict the ``training'' of the classifier $\phi$ to be done via the solution of a least-squares problem. Given the training set $\{x_i,y_i\}_{i=1}^n$, we assume that there exists some feature mapping $h : \mathbb{R}^d \to \mathbb{R}^k$ (where $k < n$). If existing linear features are not available, this could instead consist of deep features learned from a similar task---the transferability of such features is well documented \citep{Donahue, BoKernel, FeatureTransfer}---or features could be learned in an unsupervised fashion on $x_{1:n}$ (learning the features from poisoned labels could degrade performance). Given this feature mapping, let $\mX = h(x_{1:n}) \in \R^{n\times k}$ be the training point features and let $\mathbf{y} = y_{1:n}\in\{0,1\}^n$ be the labels. Our training process consists of finding the least-squares fit to the training data, i.e., we find parameters $\hat{\boldsymbol\beta} \in \mathbb{R}^k$ via the normal equation $\hat{\boldsymbol \beta} = \left (\mX^T \mX \right)^{-1} \mX^T \mathbf{y}$ and then we make a prediction on the new example via the linear function $h(x_{n+1}) \hat{\boldsymbol \beta}$.  Although it may seem odd to fit a classification task with least-squares loss, binary classification with linear regression is equivalent to Fisher's linear discriminant \citep{KernelDiscriminants} and works quite well in practice.

The real advantage of the least-squares approach is that it reduces the prediction to a linear function of $\mathbf{y}$, and thus randomizing over the labels is straightforward.  Specifically, letting
\begin{equation*}
    \boldsymbol \alpha = \mX \left (\mX^T \mX \right)^{-1} h(x_{n+1})^T,
\end{equation*}
the prediction $h(x_{n+1})\hat{\boldsymbol \beta} $ can be equivalently given by $\boldsymbol \alpha^T \mathbf{y}$ (this is effectively the kernel representation of the linear classifier).  Thus, we can simply compute $\boldsymbol \alpha$ one time and then randomly sample many different sets of labels in order to build a standard randomized smoothing bound. Further, we can pre-compute just the $\mX \left (\mX^T \mX \right)^{-1}$ term and reuse it for each test point.

\paragraph{$\ell_2$ regularization for better conditioning} It is unlikely to be the case that the training points are well-behaved for linear classification in the feature space. To address this, we instead solve an $\ell_2$ regularized version of least-squares. This is a common tool for solving systems with ill-conditioned or random design matrices \citep{RandomDesign, Regularization}. Luckily, there still exists a pre-computable closed-form solution to this problem, whereby we instead solve
\begin{equation*}
    \boldsymbol \alpha = \mX (\mX^T\mX + \lambda\mI)^{-1}h(x_{n+1})^T.
\end{equation*}
The other parts of our algorithm remain unchanged. Following results in \citet{Regularization}, we set the regularization parameter $\lambda = (1+q)\frac{\hat\sigma^2 k}{2n}\kappa(\mX^T\mX)$ for all our experiments, where $\hat\sigma^2 = \frac{\|\mathbf{y} - \mX\hat{\boldsymbol\beta}_{OLS}\|_2^2}{n-k}$ is an estimate of the variance \citep{VarianceEstimation} and $\kappa(\cdot)$ is the condition number.

\paragraph{Efficient tail bounds via the Chernoff inequality}

Even more compelling, due to the linear structure of this prediction, we can forego a sampling-based approach entirely and directly bound the tail probabilities using Chernoff bounds. Because the underlying binary prediction function $\phi$ will output the label $1$ for the test point whenever $\boldsymbol \alpha^T \mathbf{y} \geq 1/2$ and $0$ otherwise, we can derive an analytical upper bound on the probability that $g$ predicts one label or the other via the Chernoff bound. By upper bounding the probability of the \emph{opposite} prediction, we simultaneously derive a lower bound on $p$ which can be plugged in to (\ref{eq:kl-bound}) to determine the classifier's robustness. Concretely, we can upper bound the probability that the classifier outputs the label 0 by
\begin{multline}
\label{eq:chernoff-bound}
    P(\boldsymbol \alpha^T \mathbf{y} \leq 1/2) \leq \min_{t > 0} \left \{  e^{t/2} \prod_{i=1}^n E[e^{-t\boldsymbol \alpha_i y_i}]  \right \} \\= \min_{t > 0} \left \{ e^{t/2} \prod_{i=1}^n  q e^{-t\boldsymbol \alpha_i (1-y_i)} + (1-q)e^{-t\boldsymbol \alpha_i y_i} \right \}.
\end{multline}
Conversely, the probability that the classifier outputs the label 1 is upper bounded by (\ref{eq:chernoff-bound}) but evaluated at $-t$. Thus, we can solve the minimization problem unconstrained over $t$, and then let the sign of $t$ dictate which label to predict and the value of $t$ determine the bound. The objective (\ref{eq:chernoff-bound}) is log-convex in $t$ and can be easily solved by Newton's method. Note that in some cases, neither Chernoff upper bound will be less than $1/2$, meaning we cannot determine the true value of $g$. In these cases, we simply define the classifier's prediction to be determined by the sign of $t$. While we can't guarantee that this classification will match the true majority vote, our algorithm will certify
a robustness to 0 flips, so the guarantee is still valid. We avoid abstaining so as to assess our classifier's non-robust accuracy.

The key property we emphasize is that, unlike previous randomized smoothing applications, the final algorithm involves \emph{no randomness whatsoever}. Instead, the probabilities are bounded directly via the Chernoff bound, without any need for Monte Carlo approximation. Thus, the method is able to generate \emph{truly certifiable} robust predictions using approximately the same complexity as traditional predictions.

\section{Experiments}

Following \citet{KohInfluence} and \citet{SteinhardtCertified}, we perform experiments on MNIST 1/7, the IMDB review sentiment dataset \citep{imdb}, and the Dogfish binary classification challenge taken from ImageNet. We run additional experiments on multi-class MNIST and CIFAR10. For each dataset and each noise level $q$ we report the \emph{certified test set accuracy} at $r$ training label flips. That is, for each possible number of flips $r$, we plot the fraction of the test set that was both correctly classified and certified to not change under at least $r$ flips.

As mentioned, our classifier suffers an additional linear cost in the number of training points due to the kernel representation $\boldsymbol{\alpha}^T\mathbf{y}$. For most datasets there was no discernible difference in the time required to certify an input via our technique versus neural network classiﬁcation. For larger training sets such as CIFAR10, especially when doing pairwise comparisons for the multi-class case, the algorithm is embarassingly parallel; this parallelism brings runtime back in line with standard classification.

\begin{figure}[!ht]%
    \centering
    \begin{subfigure}{\linewidth}
        \includegraphics[width=\linewidth]{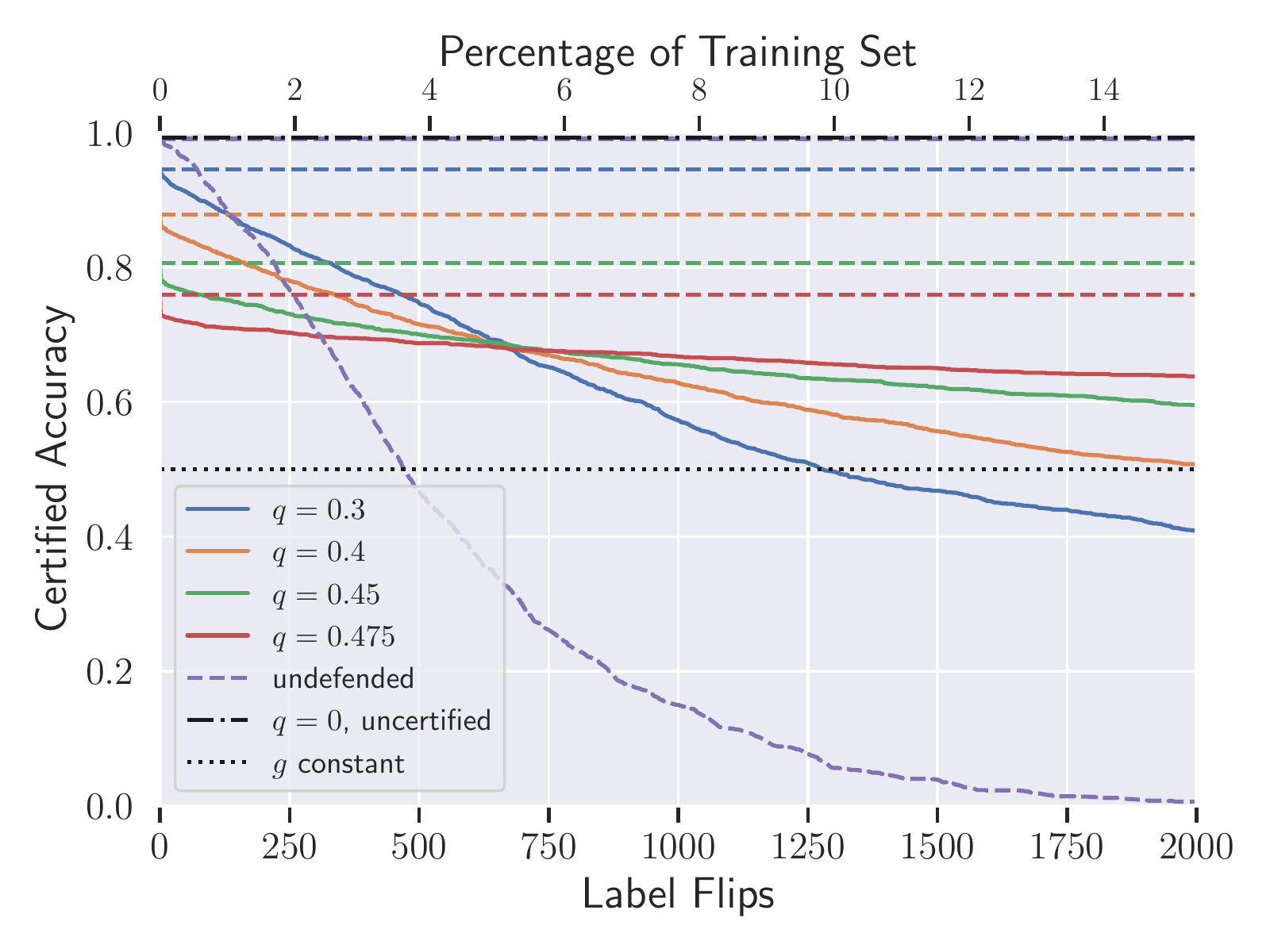}
        \caption{Binary MNIST (classes 1 and 7)}
        \label{fig:mnist-1-7}
    \end{subfigure}
    \begin{subfigure}{\linewidth}
        \includegraphics[width=\linewidth]{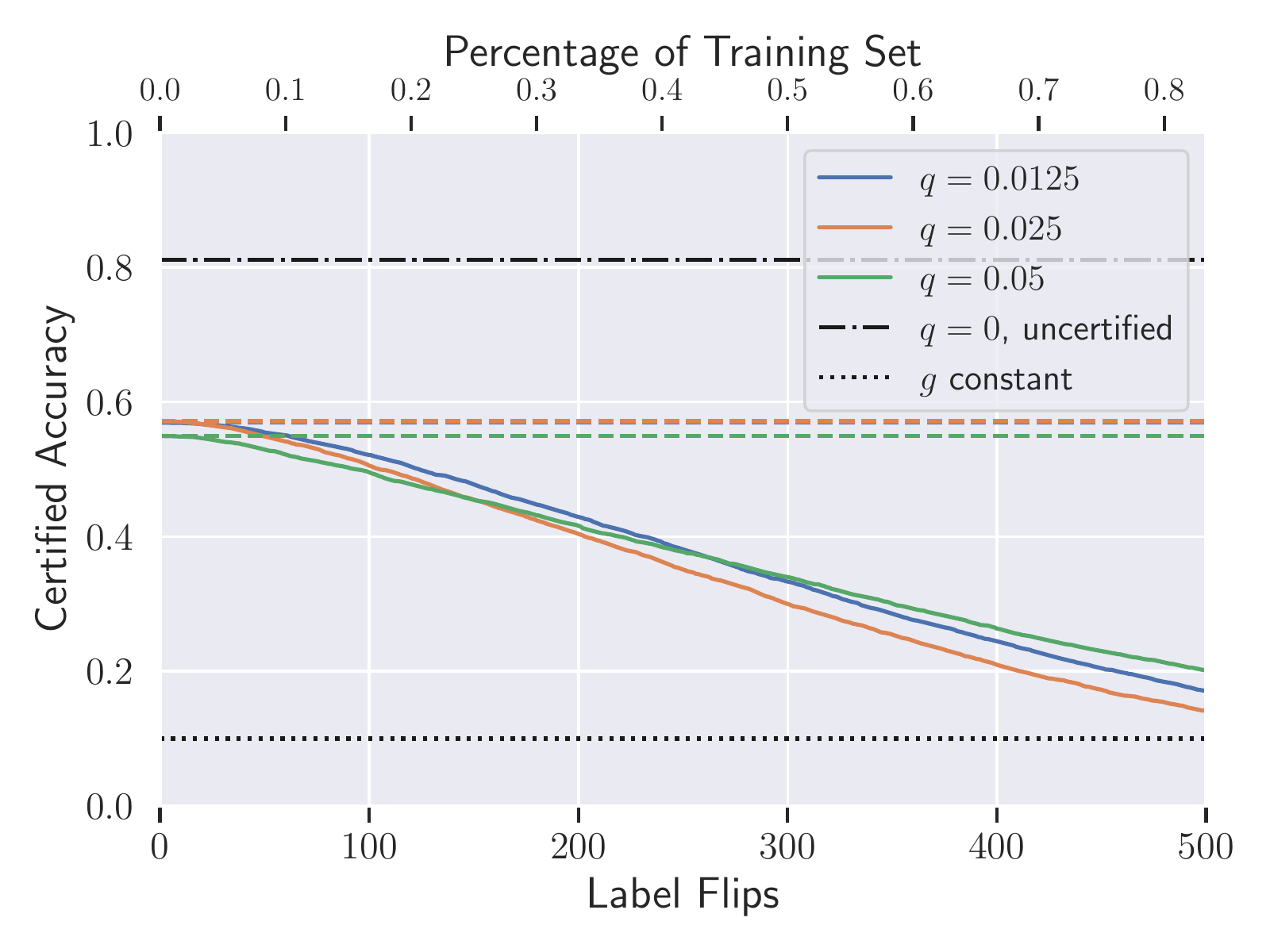}
        \caption{Full MNIST}
        \label{fig:mnist-full}
    \end{subfigure}
    \caption{MNIST 1/7 ($n=13007$, top) and full MNIST ($n=60000$, bottom) test set certified accuracy to adversarial label flips as $q$ is varied. The bottom axis represents the number of adversarial label flips to which \emph{each individual prediction} is robust, while the top axis is the same value expressed as a percentage of the training set size. The solid lines represent certified accuracy; dashed lines of the same color are the overall non-robust accuracy of each classifier. The black dotted line is the (infinitely robust) performance of a constant classifier, while the black dash-dot line is the (uncertified) performance of our classifier with no label noise.}%
    \label{fig:mnist}
\end{figure}

For binary classification, one could technically achieve a certified accuracy of 50\% at $r=\infty$ (or 10\% for MNIST or CIFAR10) by letting $g$ be constant---a constant classifier would be infinitely robust. Though not a very meaningful baseline, we include the accuracy of such a classifier in our plots (black dotted line) as a reference. We also evaluated our classifier with $q=0$ (black dash-dot line); this cannot certify robustness, but it indicates the quality of the features.

To properly justify the need for such certified defenses, and to get a sense of the scale of our certifications, we generated label-flipping attacks against the undefended binary MNIST and Dogfish models. Following previous work, the undefended models were implemented as convolutional neural networks, trained on the clean data, with all but the top layer frozen---this is equivalent to multinomial logistic regression on the learned features. For each test point we recorded how many flips were required to change the network's prediction. This number serves as an upper bound for the robustness of the network on that test point, but we note that our attacks were quite rudimentary and could almost certainly be improved upon to tighten this upper bound. Appendix \ref{appendix:attack-description} contains the details of our attack implementations. Finally, we implemented attacks on our own defense to derive an empirical upper bound and found that it reasonably tracks our lower bound. Plots and details of this attack can be found in Appendix \ref{appendix:own-attack-description}.

In all plots, the solid lines represent certified accuracy (except for the undefended classifier, which is an upper bound), while the dashed lines of the same color are the overall non-robust accuracy of each classifier.

\paragraph{Results on MNIST}

The MNIST 1/7 dataset \citep{MNIST} consists of just the classes 1 and 7, totalling 13,007 training points and 2,163 test points. We trained a simple convolutional neural network on the other eight MNIST digits to learn a 50-dimensional feature embedding and then calculated Chernoff bounds for $G$ as described in Section \ref{sect:efficient-implementation}. Figure \ref{fig:mnist-1-7} displays the certified accuracy on the test set for varying probabilities $q$. As in prior work on randomized smoothing, the noise parameter $q$ balances a trade-off; as $q$ increases, the required margin $|G - \frac{1}{2}|$ to certify a given number of flips decreases. On the other hand, this results in more noisy training labels, which reduces the margin and therefore results in lower robustness and often lower accuracy. Figure \ref{fig:mnist-full} depicts the certified accuracy for the full MNIST test set---see Appendix \ref{appendix:multi-class} for derivations of the bounds and optimization algorithm in the multi-class case. In addition to this being a significantly more difficult classification task, our classifier could not rely on features learned from other handwritten digits; instead, we extracted the top 30 components with ICA \citep{FastICA} independently of the labels. Despite the lack of fine-tuned features, our algorithm still achieves significant certified accuracy under a large number of adversarial label flips.

See Figure \ref{fig:mnist-17-regular} in the Appendix for the effect of $\ell_2$ regularization for the binary case. At a moderate cost to non-robust accuracy, the regularization results in substantially higher certified accuracy at almost all radii. We observed that regularization did not make a large difference for the multi-class case, possibly due to the inaccuracy of the residual term in the noise estimate.

\begin{figure}%
    \centering
    \includegraphics[width=\linewidth]{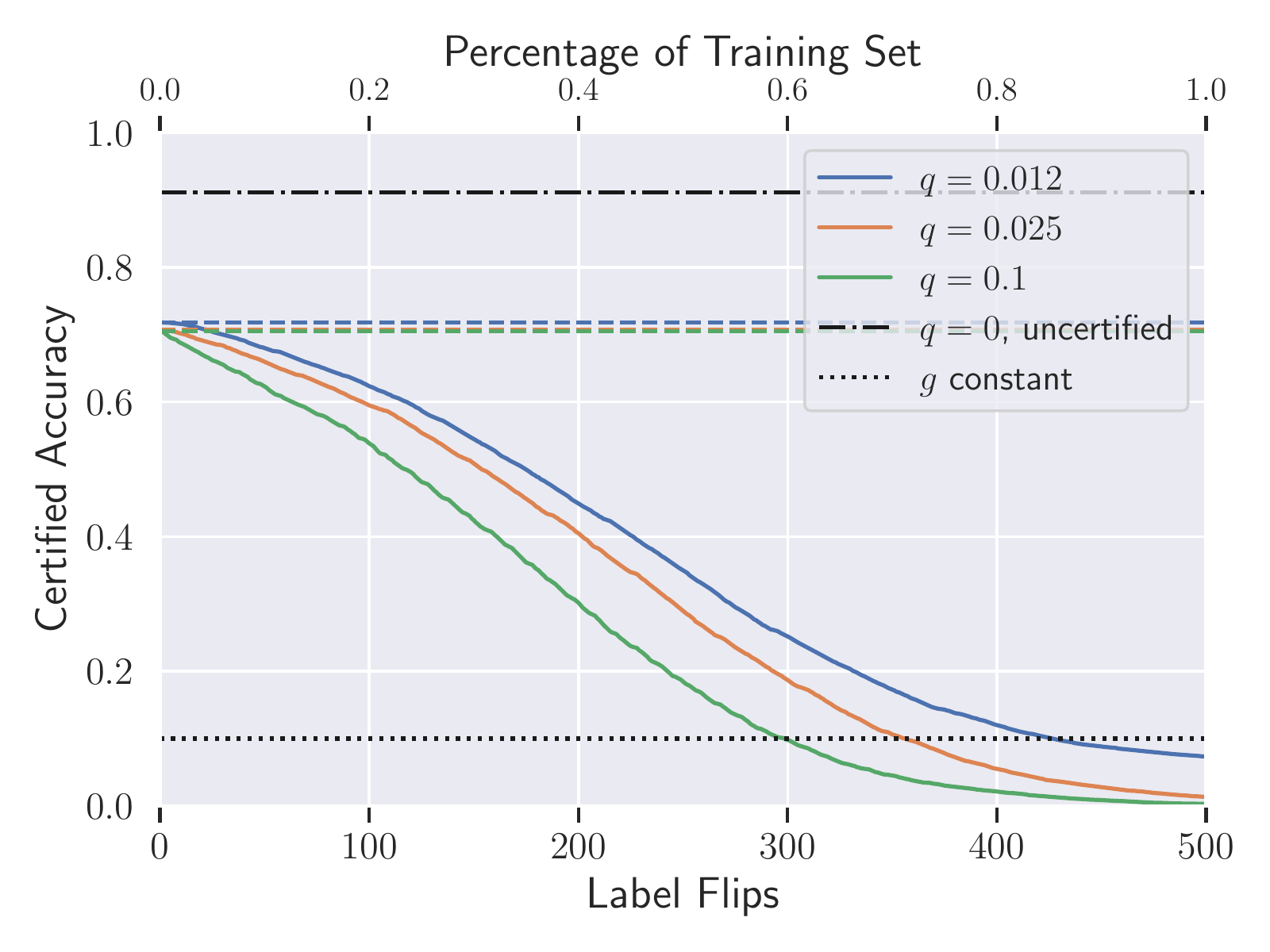}
    \caption{CIFAR10 ($n = 50000$) test set certified accuracy to adversarial label flips as $q$ is varied.}
    \label{fig:cifar}%
\end{figure}

\paragraph{Results on CIFAR10}
To further demonstrate the effectiveness of our classifier with unsupervised features, we used SimCLR \citep{SimCLR} to learn unsupervised features for CIFAR10. We used PCA to reduce the features to 128 dimensions to reduce overfitting. Figure \ref{fig:cifar} shows the results: our classifier with $q=0.12$ achieves 50\% certified accuracy up to 175 labels flips (recall there are ten classes, not two) and decays gracefully. Further, the classifier maintains better than random chance certified accuracy up to 427 label flips, which is approximately 1\% of the training set.

Because the ``votes" are changed by flipping so few labels, high values of $q$ reduce the models' predictions to almost pure chance---this means we are unable to achieve the margins necessary to certify a large number of flips. We therefore found that smaller levels of noise achieved higher certified test accuracy. This suggests that the more susceptible the original, non-robust classifier is to label flips, the lower $q$ should be set for the corresponding randomized classifier.

For much smaller values of $q$, slight differences did not decrease the non-robust accuracy---they did however have a large effect on certified robustness. This indicates that the sign of $t^\star$ is relatively stable, but the margin of $G$ is much less so. This same pattern was observed with the IMDB and Dogfish datasets. We used a high-precision arithmetic library \citep{mpmath} to achieve the necessary lower bounds, but the precision required for non-vacuous bounds grew extremely fast for $q<10^{-4}$; optimizing (\ref{eq:chernoff-bound}) quickly became too computationally expensive.

\begin{figure}%
    \centering
    \includegraphics[width=\linewidth]{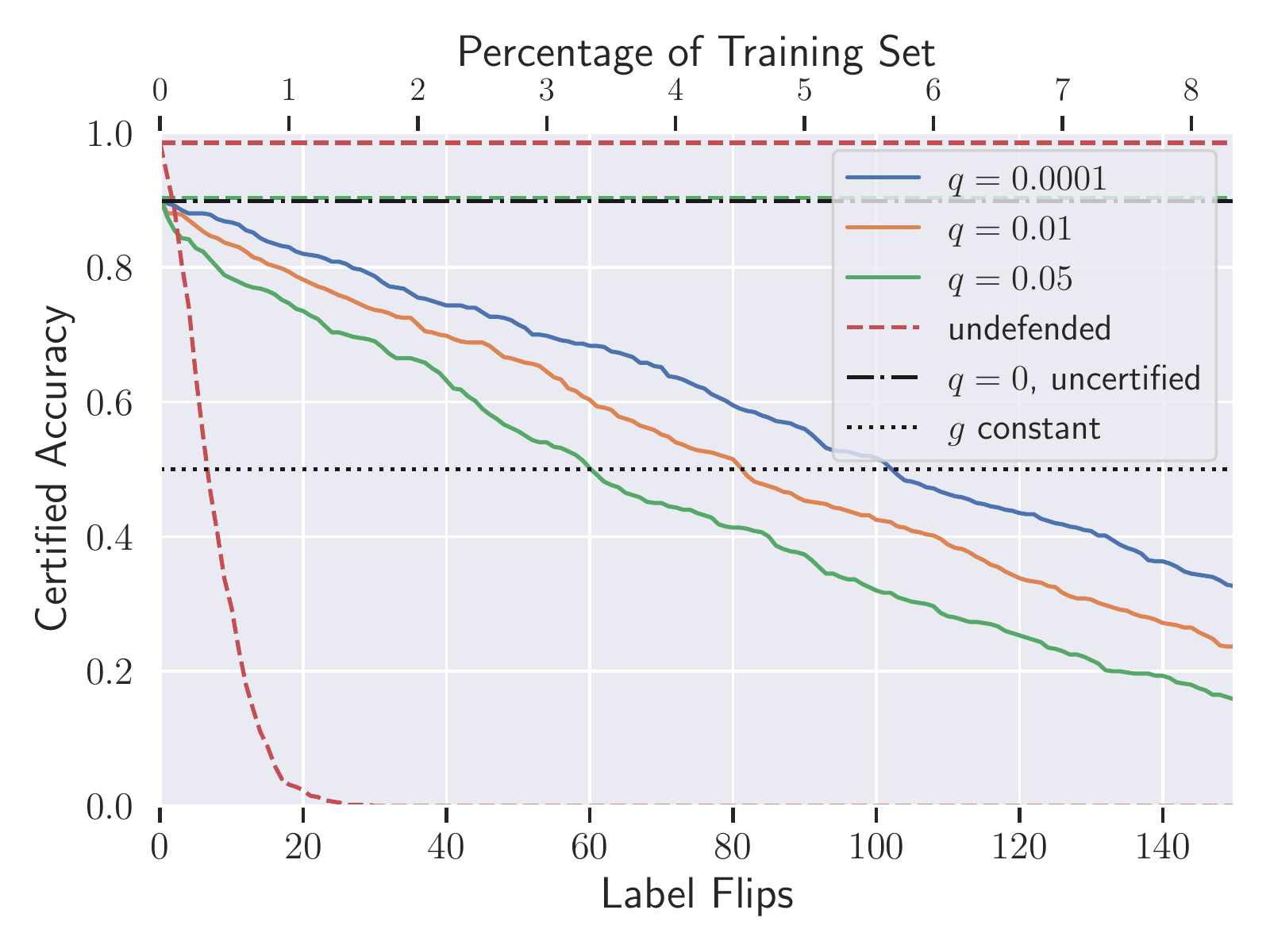}
    \caption{Dogfish ($n = 1800$) test set certified accuracy to adversarial label flips as $q$ is varied.}%
    \label{fig:dogfish}%
\end{figure}

\paragraph{Results on Dogfish}

The Dogfish dataset contains images from the ImageNet dog and fish synsets, 900 training points and 300 test points from each. We trained a ResNet-50 \citep{He_2016_CVPR} on the standard ImageNet training set but removed all images labeled dog or fish. Our pre-trained network therefore learned meaningful image features but had no features specific to either class. We again used PCA to reduce the feature space dimensionality. Figure \ref{fig:dogfish} displays the results of our poisoning attack along with our certified defense. Under the undefended model, more than 99\% of the test points can be successfully attacked with no more than 23 label flips, whereas our model with $q=10^{-4}$ can certifiably correctly classify 81.3\% of the test points under the same threat model. It would take more than four times as many flips---more than 5\% of the training set---for \emph{each test point individually} to reduce our classifier to less than 50\% certified accuracy.

Here we observe the same pattern, where reducing $q$ does not have a large effect on non-robust accuracy but does increase robustness significantly. This provides further evidence for the hypothesis that more complex datasets/classifiers are more susceptible to attacks and should be smoothed with less label noise.

Figure \ref{fig:dogfish-rand} in the Appendix shows our classifier's performance with unsupervised features. Because Dogfish is such a small dataset ($n=1800$), deep unsupervised feature learning techniques were not feasible---we instead learned overcomplete features on 16x16 image patches using RICA \citep{UnsupervisedFeatures}.

\begin{figure}
    \centering
    \includegraphics[width=\linewidth]{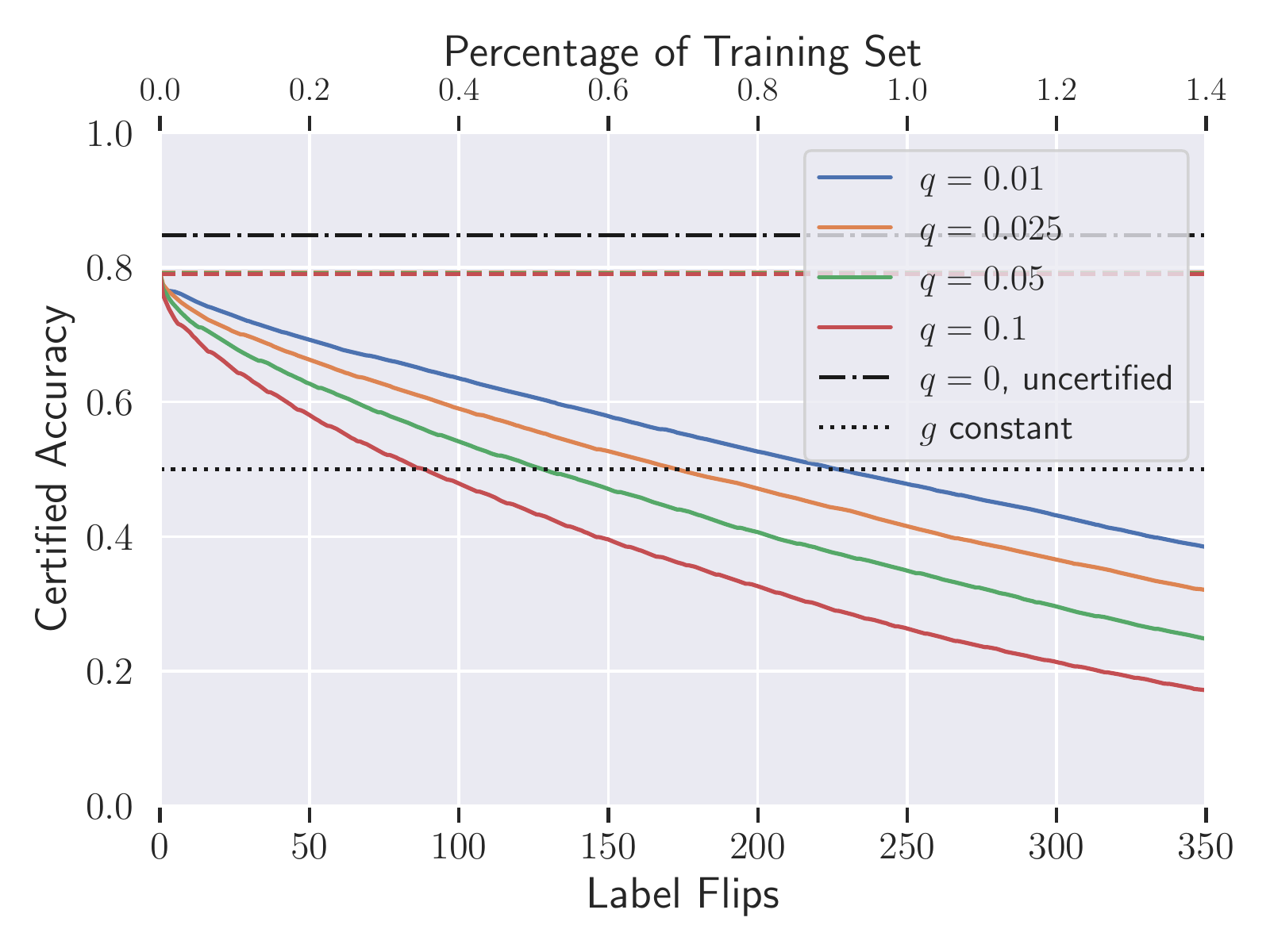}
    \caption{IMDB Review Sentiment ($n=25000$) test set certified accuracy. The non-robust accuracy slightly decreases as $q$ increases; for $q=0.01$ the non-robust accuracy is 79.11\%, while for $q=0.1$ it is 78.96\%.}
    \label{fig:imdb}
\end{figure}

\paragraph{Results on IMDB}
Figure \ref{fig:imdb} plots the result of our randomized smoothing procedure on the IMDB review sentiment dataset. This dataset contains 25,000 training examples and 25,000 test examples, evenly split between ``positive" and ``negative". To extract the features we applied the Google News pre-trained Word2Vec to all the words in each review and averaged them. This feature embedding is considerably noisier than that of an image dataset, as most of the words in a review are irrelevant to sentiment classification. Indeed, \citet{SteinhardtCertified} also found that the IMDB dataset was much more susceptible to adversarial corruption than images when using bag-of-words features. Consistent with this, we found smaller levels of noise  resulted in larger certified accuracy. We expect significant improvements could be made with a more refined choice of feature embedding.

\section{Conclusion}
In this work we presented a unifying view of randomized smoothing, which borrows from the literature of differential privacy in order to provide black-box certificates of robustness. Based on the observation that this framework is applicable more broadly than just defenses to adversarial examples, we used it to derive a framework for certified defenses against arbitrary data poisoning attacks---we dub such defenses ``pointwise" because they provide a certificate of robustness for each test point.

We next implemented this protocol as a specific classifier which is robust to a strong class of label-flipping attacks, where an adversary can flip labels to target each test point individually. This contrasts with previous data poisoning defenses which have typically only considered an adversary who wishes to degrade the classifier's accuracy on the test distribution as a whole. Finally, we offered a tractable algorithm for evaluating this classifier which, despite being rooted in randomization, can be computed with no Monte Carlo sampling whatsoever, resulting in a truly certifiably robust classifier. This work represents the first classification algorithm that is pointwise-certifiably robust to \emph{any type} of data poisoning attack; we anticipate many possible new directions within this framework.

A particular strength of this framework is when we specifically care about robustly classifying each input individually. Compared to traditional robust classification, this technique is superior for determining who receives a coveted resource (a loan, parole, etc.) or for making some other sensitive classification, as it provides a guarantee for each individual. Other works only ensure that they correctly classify some fraction $p$ of the population, which is often not acceptable as that still leaves the $1-p$ fraction who could be misclassified, with no indication of which ones belong to the test set.

There are several avenues for improvements to this line of work. Most immediately, our protocol could be implemented with other types of smoothing distributions applied to the training data, such as randomizing over the input data or features, to derive specific algorithms that are pointwise-certifiably robust to other types of data poisoning attacks. Additionally, the method for learning the input features in an unsupervised, semi-supervised, or self-supervised manner could be improved. Finally, we hope that our defense to this threat model will inspire the development of more powerful (e.g., pointwise) train-time attacks, against which future defenses can be evaluated.

\section*{Acknowledgements}
We thank Guang-He Lee for sharing his code on discrete robustness certificates, Adarsh Prasad for several helpful sources on robust linear regression and classification, and Alnur Ali for pointing us to \citet{VarianceEstimation} for help choosing the appropriate regularization term. We are grateful to Jeremy Cohen and Adarsh Prasad for helpful discussions and reviewing drafts of this work. E. R. and P. R. acknowledge the support of DARPA via HR00112020006.

\bibliography{icml2020}
\bibliographystyle{icml2020}

\appendix
\onecolumn
\counterwithin{figure}{section}

\section{Generic Randomized Smoothing Algorithm}
\label{appendix:generic-randomized-smoothing}
\begin{algorithm}[H]
\caption{Generic randomized smoothing procedure}
\label{alg:generic-randomized-smoothing}
\begin{algorithmic}
\STATE \textbf{Input:} function $\phi : \mathcal{X} \to \{0,1\}$, number of samples $N$, smoothing distribution $\mu$, test point to predict $x_0$, failure probability $\delta > 0$.
\FOR{$i = 1,\ldots,N$}
\STATE Sample $x_i \sim \mu(x_0)$ and compute $y_i = \phi(x_i)$.
\ENDFOR
\STATE Compute approximate smoothed output
\begin{equation*}
    \hat{g}(\mu,\phi) = \mathbf{1}\left\{\left(\frac{1}{N}\sum_{i=1}^N y_i\right) \geq \frac{1}{2} \right\}.
\end{equation*}
\STATE Compute bound $\hat{G}(\mu,\phi)$ such that with probability $\geq 1-\delta$
\begin{equation*}
    \hat{G}(\mu,\phi) \left\{ \begin{array}{ll}\leq G(\mu, \phi) & \mbox{if } \hat{g}(\mu, \phi) = 1 \\
    \geq G(\mu, \phi) & \mbox{if } \hat{g}(\mu,\phi) = 0. \end{array} \right .
\end{equation*}

\STATE \textbf{Output:} Prediction $\hat{g}(\mu,\phi)$ and probability bound $\hat{G}(\mu,\phi)$, or abstention if $\hat{g}(\mu,\phi) \neq \textrm{sign}(\hat G(\mu, \phi)-\frac{1}{2}).$
\end{algorithmic}
\end{algorithm}

\section{The Multi-Class Setting}
\label{appendix:multi-class}

Although the notation and algorithms are slightly more complex, all the methods we have discussed in the main paper can be extended to the multi-class setting.  In this case, we consider a class label $y \in \{1,\ldots,K\}$, and we again seek some smoothed prediction such that the classifier's prediction on a new point will not change with some number $r$ flips of the labels in the training set.

\subsection{Randomized smoothing in the multi-class case}
We here extend our notation to the case of more than two classes. Recall our original definition of $G$,
\begin{equation*}
    G(\mu, \phi) = \mathbf{E}_{x\sim\mu}[\phi(x)]=\int_{\mathcal{X}}\mu(x)\phi(x)dx,
\end{equation*}
where $\phi:\mathcal{X}\to\{0,1\}$. More generally, consider a classifier $\phi:\mathcal{X}\to[K]$, outputting the index of one of $K$ classes. Under this formulation, for a given class $c\in[K]$, we have
\begin{equation*}
    G(\mu,\phi,c) = \mathbf{E}_{x\sim\mu}[\phi_c(x)] = \int_{\mathcal{X}}\mu(x)\phi_c(x)dx,
\end{equation*}
where $\phi_c(x) = \mathbf{1}\left\{\phi(x) = c\right\}$ is the indicator function for if $\phi(x)$ outputs the class $c$. In this case, the hard threshold $g$ is evaluated by returning the class with the highest probability. That is,
\begin{equation*}
    g(\mu, \phi) = \arg\max_c G(\mu, \phi, c).
\end{equation*}

\subsection{Linearization and Chernoff bound approach for the multi-class case}

Using the same linearization approach as in the binary case, we can formulate an analogous approach which forgoes the need to actually perform random sampling at all and instead directly bounds the randomized classifier using the Chernoff bound.

Adopting the same notation as in the main text, the equivalent least-squares classifier for the multi-class setting finds some set of weights
\begin{equation*}
    \hat{\boldsymbol \beta} = \left (\mX^T \mX \right)^{-1} \mX^T \mY
\end{equation*}
where $\mY \in \{0,1\}^{n \times K}$ is a binary matrix with each row equal to a one-hot encoding of the class label (note that the resulting $\hat{\boldsymbol \beta} \in \mathbb{R}^{k \times K}$ is now a matrix, and we let $\hat{\boldsymbol \beta_i}$ refer to the $i$th column).  At prediction time, the predicted class of some new point $x_{n+1}$ is simply given by the prediction with the highest value, i.e.,
\begin{equation*}
    \hat{y}_{n+1} = \arg\max_{i} \hat{\boldsymbol \beta_i}^T h(x_{n+1}).
\end{equation*}
Alternatively, following the same logic as in the binary case, this same prediction can be written in terms of the $\boldsymbol \alpha$ variable as 
\begin{equation*}
    \hat{y}_{n+1} = \arg\max_{i} \boldsymbol \alpha^T \mY_i
\end{equation*}
where $\mY_i$ denotes the $i$th column of $\mY_i$.

In our randomized smoothing setting, we again propose to flip the class of any label with probability $q$, selecting an alternative label uniformly at random from the remaining $K-1$ labels.  Assuming that the predicted class label is $i$, we wish to bound the probability that
\begin{equation*}
    P(\boldsymbol \alpha^T \mY_i < \boldsymbol \alpha^T \mY_{i'})
\end{equation*}
for all alternative classes $i'$.  By the Chernoff bound, we have that
\begin{equation*}
\begin{split}
    \log P(\boldsymbol \alpha^T \mY_i < \boldsymbol \alpha^T \mY_{i'})
    & = \log P(\boldsymbol \alpha^T(\mY_i - \mY_{i'}) \leq 0) \\
    & \leq \min_{t> 0} \left \{ \sum_{j=1}^n \log \mathbf{E}\left [e^{-t \boldsymbol \alpha_j (\mY_{ji} - \mY_{ji'})} \right] \right \}.
    \end{split}    
\end{equation*}
The random variable $\mY_{ji} - \mY_{ji'}$ takes on three different distributions depending on if $y_j = i$, if $y_j = i'$, or if $y_j \neq i$ and $y_j \neq i'$.  Specifically, this variable can take on the terms $+1,0,-1$ with the associated probabilities
\begin{equation*}
    \begin{split}
    P(\mY_{ji} - \mY_{ji'} = +1) & = 
    \left \{ \begin{array}{ll} 
    1-q & \mbox{if } y_j = i, \\
    q/(K-1) & \mbox{otherwise.} \end{array} \right . \\
    P(\mY_{ji} - \mY_{ji'} = -1) & = 
    \left \{ \begin{array}{ll} 
    1-q & \mbox{if } y_j = i', \\
    q/(K-1) & \mbox{otherwise.} \end{array} \right . \\ 
    P(\mY_{ji} - \mY_{ji'} = 0) & = 
    \left \{ \begin{array}{ll} 
    q(K-2)/(K-1) & \mbox{if } y_j=i \mbox{ or } y_j=i', \\
    1-2q/(K-1) & \mbox{otherwise.} \end{array} \right . \\ 
    \end{split}
\end{equation*}

Combining these cases directly into the Chernoff bound gives
\begin{equation*}
\begin{split}
    \log P(\boldsymbol \alpha^T \mY_i < \boldsymbol \alpha^T \mY_{i'}) \leq \min_{t> 0} \biggl \{ &
    \sum_{j:y_j = i} \log \left ((1-q)e^{-t \boldsymbol \alpha_j} + q\frac{K-2}{K-1} + \frac{q}{K-1} e^{t\boldsymbol \alpha_j} \right ) + \\
    & \sum_{j:y_j = i'} \log \left (\frac{q}{K-1}e^{-t \boldsymbol \alpha_j} + q\frac{K-2}{K-1} + (1-q) e^{t\boldsymbol \alpha_j} \right ) + \\
    & \sum_{j:y_j \neq i, y_j \neq i'} \log \left (\frac{q}{K-1}e^{-t \boldsymbol \alpha_j} + 1- 2\frac{q}{K-1} + \frac{q}{K-1} e^{t\boldsymbol \alpha_j} \right ) \biggr \}.
     \end{split}    
\end{equation*}
Again, this problem is convex in $t$, and so can be solved efficiently using Newton's method.  And again since the reverse case can be computed via the same expression we can similarly optimize this in an unconstrained fashion.  Specifically, we can do this for every pair of classes $i$ and $i'$, and return the $i$ which gives the smallest lower bound for the worst-case choice of $i'$.

\subsection{KL Divergence Bound}
\label{appendix:kl-bound-derivation}
To compute actual certification radii, we will derive the KL divergence bound for the the case of $K$ classes.  Let $\mu, \rho$ be defined as in Section \ref{sec:kl-bound}, except that as in the previous section when a label is flipped with probability $q$ it is changed to one of the other $K-1$ classes uniformly at random. Let $\mu_i$ and $\rho_i$ refer to the independent measures on each dimension which collectively make up the factorized distributions $\mu$ and $\rho$ (i.e., $\mu(x) = \prod_{i=1}^d \mu_i(x)$). Further, let $Y_1^i$ be the $i^{th}$ element of $Y_1$, meaning it is the ``original" label which may or may not be flipped when sampling from $\mu$. First noting that each dimension of the distributions $\mu$ and $\rho$ are independent, we have
\begin{align*}
    \mathrm{KL}\infdivx{\rho}{\mu} &= \sum_{i=1}^n \mathrm{KL}\infdivx{\rho_i}{\mu_i}\\
    &= \sum_{i:\rho_i\neq\mu_i} \mathrm{KL}\infdivx{\rho_i}{\mu_i}\\
    &= r\left( \sum_{j=1}^K \rho_i(j)\log\left(\frac{\rho_i(j)}{\mu_i(j)}\right) \right) \\
    &= r\left( \rho_i(Y_1^i)\log\left(\frac{\rho_i(Y_1^i)}{\mu_i(Y_1^i)}\right) + \rho_i(Y_2^i)\log\left(\frac{\rho_i(Y_2^i)}{\mu_i(Y_2^i)}\right)  \right)\\
    &= r\left( (1-q)\log\left(\frac{1-q}{\frac{q}{K-1}}\right) + \frac{q}{K-1}\log\left(\frac{\frac{q}{K-1}}{1-q}\right)\right) \\
    &= r\left(1-\frac{Kq}{K-1}\right) \log\left(\frac{(1-q)(K-1)}{q}\right).
\end{align*}

Plugging in the robustness guarantee (\ref{eq:generic-bound}), we have that $g(\mu, \phi) = g(\rho, \phi)$ so long as
\begin{equation*}
    r \le \frac{\log(4p(1-p))}{2(1-\frac{Kq}{K-1})\log\left(\frac{q}{(1-q)(K-1)}\right)}.
\end{equation*}

Setting $K=2$ recovers the divergence term (\ref{eq:kl-divergence}) and the bound (\ref{eq:kl-bound}).

\section{Description of Label-Flipping Attacks}

\subsection{Attacks on Undefended Classifiers}
\label{appendix:attack-description}
Due to the dearth of existing work on label-flipping attacks for deep networks, our attacks on MNIST and Dogfish were quite straightforward; we expect significant improvements could be made to tighten this upper bound.

For Dogfish, we used a pre-trained Inception network \citep{Szegedy_2016_CVPR} to evaluate the influence of each training point with respect to the loss of each test point \citep{KohInfluence}. As in prior work, we froze all but the top layer of the network for retraining. Once we obtained the most influential points, we flipped the first one and recomputed approximate influence using only the top layer for efficiency. After each flip, we recorded which points were classified differently and maintained for each test point the successful attack which required the fewest flips. When this was finished, we also tried the reverse of each attack to see if any of them could be achieved with even fewer flips.

For MNIST we implemented two similar attacks and kept the best attack for each test point. The first attack simply ordered training labels by their $\ell_2$ distance from the test point in feature space, as a proxy for influence. We then tried flipping these one at a time until the prediction changed, and we also tried the reverse. The second attack was essentially the same as the Dogfish attack, ordering the test points by influence. To calculate influence we again assumed a frozen feature map; specifically, using the same notation as \citet{KohInfluence}, the influence of flipping the label of a training point $z=(x,y)$ to $z^- = (x,1-y)$ on the loss at the test point $z_{\text{test}}$ is:
\begin{equation*}
\begin{split}
    \frac{dL(z_{\text{test}}, \hat\theta_{\epsilon,z^-,-z})}{d\epsilon} &= \nabla_\theta L(z_{\text{test}},\hat\theta)^T \frac{d\hat\theta_{\epsilon,z^-,-z}}{d\epsilon} \\
    &\approx -\nabla_\theta L(z_{\text{test}},\hat\theta)^T H_{\hat\theta}^{-1}\left( \nabla_\theta L(z^-,\hat\theta)-\nabla_\theta L(z,\hat\theta)\right).
\end{split}
\end{equation*}

For logistic regression these values can easily be computed in closed form.

\subsection{Attacks on Our Classifier}
\label{appendix:own-attack-description}
Recall that our theoretical classifier outputs a prediction based on $P(\boldsymbol{\alpha}^T \mathbf{y}\geq 1/2)$, where the randomness is over the label flips of $\mathbf{y}$. More specifically, the classifier's output is based on a weighted majority vote of ``sub-classifiers", each of which is a simple linear classifier which outputs $\mathbf 1\{\boldsymbol \alpha^T \mathbf{\hat y} \geq 1/2\}$ for its own labels $\mathbf{\hat y}$. The sub-classifier's vote is weighted by its probability under the smoothing distribution, which depends only on $\|\mathbf{y}-\mathbf{\hat y}\|_0$ (and is monotonically decreasing in this value). It is clear that the optimal attack to reduce $P(\boldsymbol{\alpha}^T \mathbf{y}\geq 1/2)$ is to flip the labels which will push the inner product $\boldsymbol{\alpha}^T \mathbf{y}$ as much as possible towards the incorrect label: flipping labels by their change to the inner product will add weight to the votes of the most overall number of incorrect sub-classifiers, pushing our smoothed classifier to be incorrect.

Here we make a subtle distinction: while this attack is optimal for the purpose of reducing $P(\boldsymbol{\alpha}^T\mathbf{y}\geq 1/2)$, it is \emph{not necessarily optimal} against our classifier, even though this probability represents how our classifier (theoretically) makes a prediction. This is because in practice, we never actually compute $P(\boldsymbol{\alpha}^T\mathbf{y}\geq 1/2)$. Instead, recall from \eqref{eq:chernoff-bound} that we use the Chernoff inequality to tightly bound this probability. Thus, while the attack described above is optimal for reducing the \emph{true} probability (and therefore the theoretical robustness), it is technically possible that a different attack would cause a looser Chernoff bound, more effectively reducing our bound on the probability. In essence, our attack is optimal for modifying the LHS of \eqref{eq:chernoff-bound}, but not necessarily the RHS, which is ultimately how our classifier \emph{actually} makes predictions.

\begin{figure}
\centering
\begin{subfigure}{.5\textwidth}
  \centering
  \includegraphics[width=\linewidth]{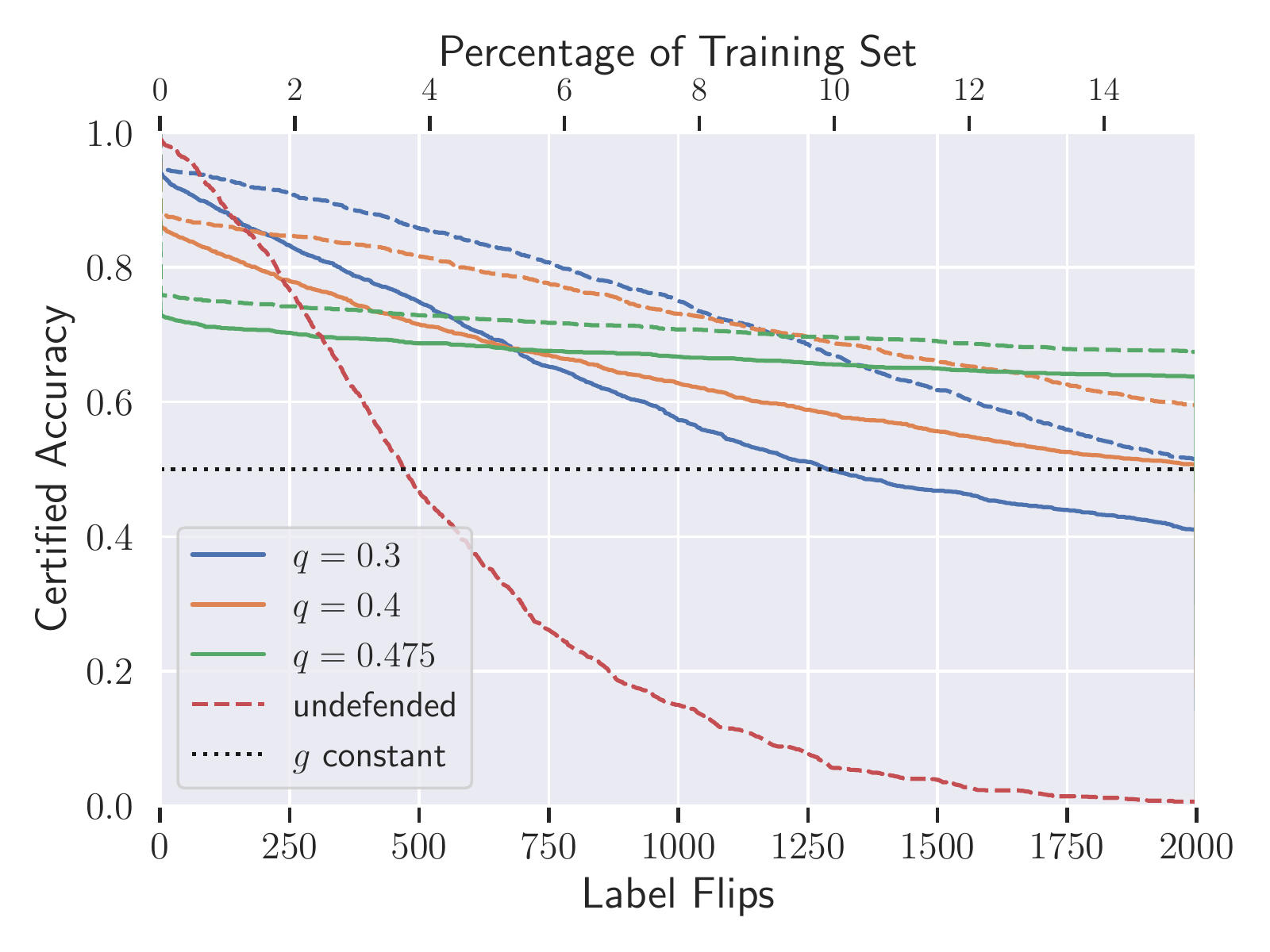}
  \caption{Binary MNIST (classes 1 and 7)}
  \label{fig:mnist-own-attack}
\end{subfigure}%
\begin{subfigure}{.5\textwidth}
  \centering
  \includegraphics[width=\linewidth]{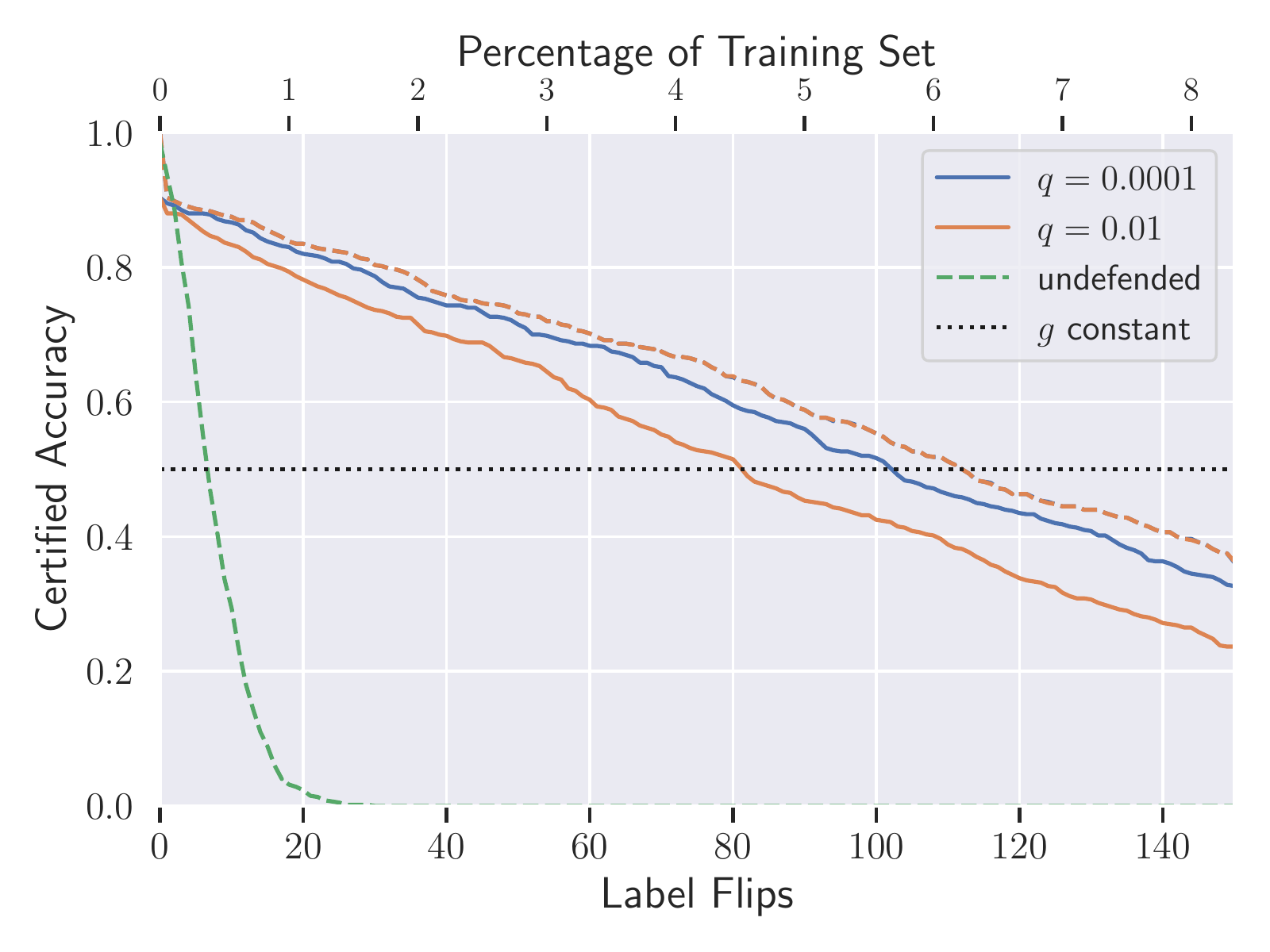}
  \caption{Dogfish}
  \label{fig:dogfish-own-attack}
\end{subfigure}
\caption{MNIST 1/7 and Dogfish certified lower bounds (solid) compared to empirical upper bounds (dashed) of our classifier and the undefended classifier. Our classifier's upper and lower bounds are reasonably close, and they get closer as $q$ decreases. The gap is due to the potential looseness of the Chernoff bound, though in practice we would expect the true robustness to be closer to the upper bound.}
\end{figure}

With that said, the existence of an attack which causes the Chernoff bound to return a particularly sub-optimal bound seems debatable. So, while we present these results as an empirical upper bound, we believe it would not be inappropriate to also view them as an \emph{approximate} lower bound. Of course, the actual lower bound returned by our classifier is still a genuinely guaranteed certificate. Figure \ref{fig:mnist-own-attack} displays the result of our attack on MNIST 1/7, with the undefended classifier for comparison. Observe that the empirical upper bounds (dashed lines) track the guaranteed lower bounds (solid lines) reasonably closely. The gap is under 10\% accuracy and shrinks as the noise $q$ decreases. Further, this empirical robust accuracy outperforms the undefended classifier's empirical robust accuracy by an even larger margin. Figure \ref{fig:dogfish-own-attack} presents the same results on the Dogfish dataset. Our empirical attacks had very similar success rates for all values of $q$, so we only plot two values along with the undefended classifier. We again observe a tight correspondence between upper and lower bounds which gets tighter with smaller $q$.

\section{Additional Tables of Results}
To supplement the line plots, for each dataset and noise parameter we present here precise certified test set accuracy at specific numbers of label flips. When available, for comparison we also provide the undefended classifier's empirical accuracy when subjected to our label-flipping attack as detailed in Section \ref{appendix:attack-description}. For each number of label flips, the noise hyperparameter setting which results in the highest certified accuracy is in bold.

\begin{table}[!ht]
    \centering
    \begin{tabular}{|c|c|c|c|c|c|c|c|}
        \hline
        \multicolumn{8}{|c|}{\textbf{MNIST 1/7} ($n=13007$, 2 classes)} \\
        \hline
        & \multicolumn{7}{|c|}{\textbf{Number of Label Flips}} \\
        \hline
        \textbf{Noise $\boldsymbol{q \downarrow}$} &  1 & 10 & 100 & 500 & 1000 & 1500 & 2000 \\
        \hline 
        (undefended) & (.9903) & (.9815) & (.9163) & (.4674) & (.1503) & (.0388) & (.0065) \\
        0.3   & \textbf{.9399} & \textbf{.9320} & \textbf{.8918} & \textbf{.7470} & .5726 & .4681 & .4089 \\
        0.4   & .8659 & .8571 & .8248 & .7152 & .6283 & .5566 & .5072 \\
        0.45  & .7855 & .7767 & .7540 & .7004 & .6556 & .6218 & .5950 \\
        0.475 & .7294 & .7262 & .7118 & .6873 & \textbf{.6674} & \textbf{.6503} & \textbf{.6378} \\
        \hline
    \end{tabular}
    \caption{Certified test set accuracy on MNIST 1/7 (Figure \ref{fig:mnist-1-7}), with the undefended classifier's empirical robust accuracy for comparison. Random guessing or a constant classifier would attain 50\% accuracy.}
\end{table}

\begin{table}[!ht]
    \centering
    \begin{tabular}{|c|c|c|c|c|c|c|c|}
        \hline
        \multicolumn{8}{|c|}{\textbf{Full MNIST} ($n=60000$, 10 classes)} \\
        \hline
        & \multicolumn{7}{|c|}{\textbf{Number of Label Flips}} \\
        \hline
        \textbf{Noise $\boldsymbol{q \downarrow}$} & 1 & 10 & 100 & 200 & 300 & 400 & 500  \\
        \hline 
        0.0125 & .5693 & .5689 & \textbf{.5212} & \textbf{.4292} & .3333 & .2446 & .1706 \\
        0.025  & \textbf{.5713} & \textbf{.5701} & .5053 & .4040 & .2999 & .2096 & .1407 \\
        0.05   & .5495 & .5486 & .4954 & .4160 & \textbf{.3400} & \textbf{.2633} & \textbf{.2012} \\
        \hline
    \end{tabular}
    \caption{Certified test set accuracy on Full MNIST (Figure \ref{fig:mnist-full}). Random guessing or a constant classifier would attain 10\% accuracy.}
\end{table}

\begin{table}[!ht]
    \centering
    \begin{tabular}{|c|c|c|c|c|c|c|c|}
        \hline
        \multicolumn{8}{|c|}{\textbf{CIFAR10} ($n=50000$, 10 classes)} \\
        \hline
        & \multicolumn{7}{|c|}{\textbf{Number of Label Flips}} \\
        \hline
        \textbf{Noise $\boldsymbol{q \downarrow}$} & 1 & 10 & 50 & 100 & 200 & 300 & 400  \\
        \hline
        0.012 & \textbf{.7180} & \textbf{.7158} & \textbf{.6800} & \textbf{.6234} & \textbf{.4493} & \textbf{.2520} & \textbf{.1201} \\
        0.025 & .7068 & .7017 & .6597 & .5949 & .4051 & .1870 & .0548 \\
        0.1   & .7040 & .6876 & .6230 & .5384 & .3019 & .0981 & .0213 \\
        \hline
    \end{tabular}
    \caption{Certified test set accuracy on CIFAR10 (Figure \ref{fig:cifar}). Random guessing or a constant classifier would attain 10\% accuracy.}
\end{table}

\begin{table}[!ht]
    \centering
    \begin{tabular}{|c|c|c|c|c|c|c|c|}
        \hline
        \multicolumn{8}{|c|}{\textbf{Dogfish} ($n=1800$, 2 classes)} \\
        \hline
        & \multicolumn{7}{|c|}{\textbf{Number of Label Flips}} \\
        \hline
        \textbf{Noise $\boldsymbol{q \downarrow}$} & 1 & 10 & 25 & 50 & 75 & 100 & 150  \\
        \hline 
        (undefended) & (.9367) & (.2933) & (.0050) & (.0000) & (.0000) & (.0000) & (.0000) \\
        0.0001 & .8950 & \textbf{.8667} & \textbf{.8083} & \textbf{.7150} & \textbf{.6233} & \textbf{.5167} & \textbf{.3267} \\
        0.001  & .8950 & .8550 & .7967 & .6967 & .5917 & .4750 & .3017 \\
        0.01   & .8800 & .8333 & .7583 & .6617 & .5283 & .4250 & .2367 \\
        0.05   & \textbf{.9367} & .7833 & .7033 & .5567 & .4350 & .3200 & .1583 \\
        \hline
    \end{tabular}
    \caption{Certified test set accuracy on Dogfish (Figure \ref{fig:dogfish}), with the undefended classifier's empirical robust accuracy for comparison. Random guessing or a constant classifier would attain 50\% accuracy.}
\end{table}

\begin{table}[!ht]
    \centering
    \begin{tabular}{|c|c|c|c|c|c|c|c|}
        \hline
        \multicolumn{8}{|c|}{\textbf{IMDB Sentiment Analysis} ($n=25000$, 2 classes)} \\
        \hline
        & \multicolumn{7}{|c|}{\textbf{Number of Label Flips}} \\
        \hline
        \textbf{Noise $\boldsymbol{q \downarrow}$} & 1 & 10 & 25 & 50 & 100 & 200 & 300  \\
        \hline
        0.01  & .6275 & .5980 & .5882 & .5686 & \textbf{.5392} & \textbf{.4804} & \textbf{.4412} \\
        0.025 & .6364 & .6154 & .5944 & .5594 & .5105 & .4406 & .3287 \\
        0.05  & .5878 & .5344 & .5038 & .4656 & .4160 & .3206 & .2519 \\
        0.1   & \textbf{.7585} & \textbf{.7034} & \textbf{.6469} & \textbf{.5806} & .4788 & .3263 & .2135 \\
        \hline
    \end{tabular}
    \caption{Certified test set accuracy on the IMDB Sentiment Analysis dataset (Figure \ref{fig:imdb}). Random guessing or a constant classifier would attain 50\% accuracy.}
\end{table}

\newpage

\section{Additional Plots}
\label{appendix:additional-plots}
\begin{figure}[!ht]
    \centering
    \includegraphics[width=0.5\linewidth]{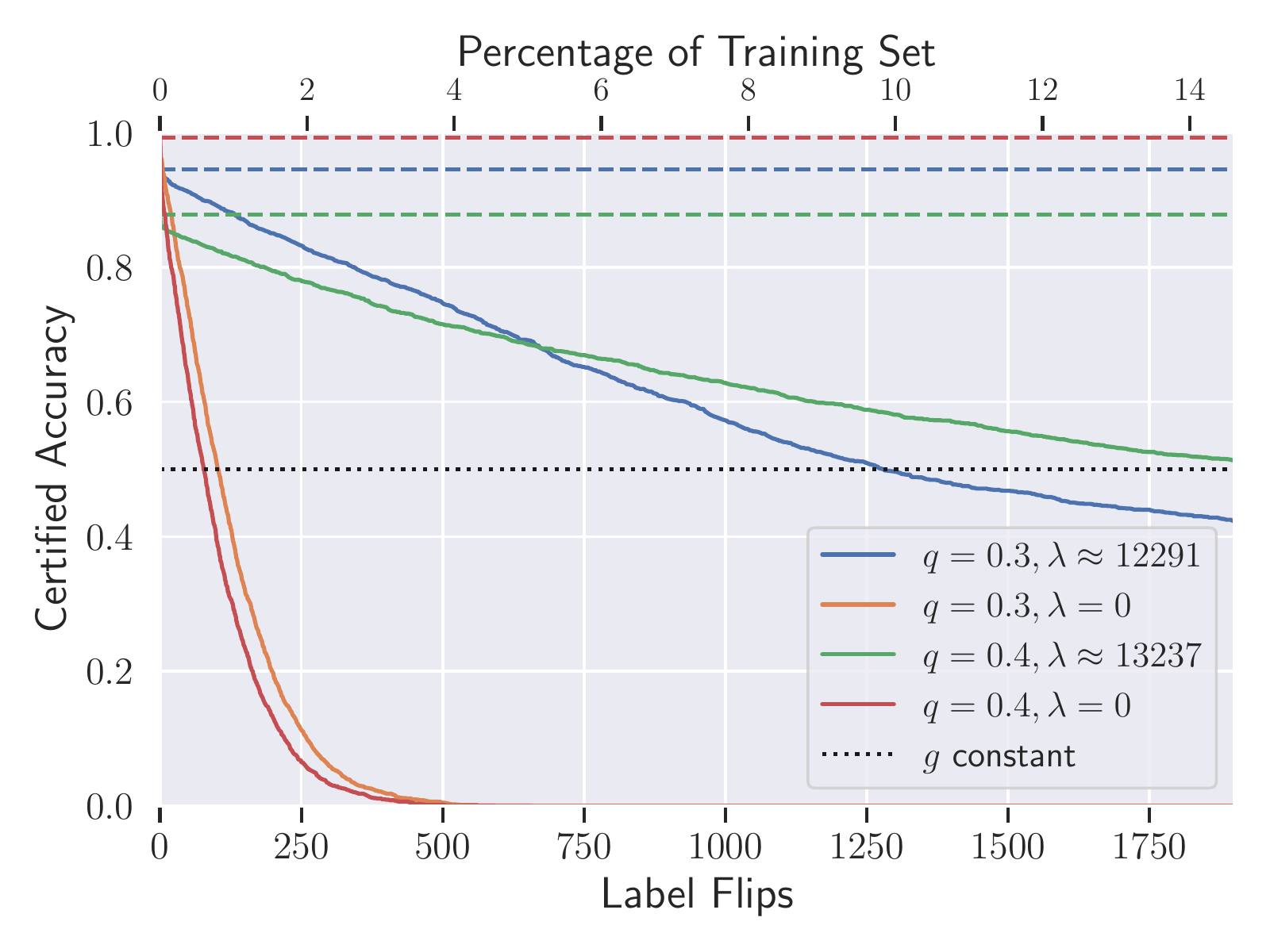}
    \caption{MNIST 1/7 test set certified accuracy with and without $\ell_2$ regularization in the computation of $\boldsymbol \alpha$. Note that the unregularized solution achieves almost 100\% non-robust accuracy, but certifies significantly lower robustness. This implies that the ``training" process is not robust enough to label noise, hence the lower margin by the ensemble. In comparison, the regularized solution achieves significantly higher margins, at a slight cost in overall accuracy.}
    \label{fig:mnist-17-regular}
\end{figure}

\begin{figure} [h]
    \centering
    \includegraphics[width=0.5\linewidth]{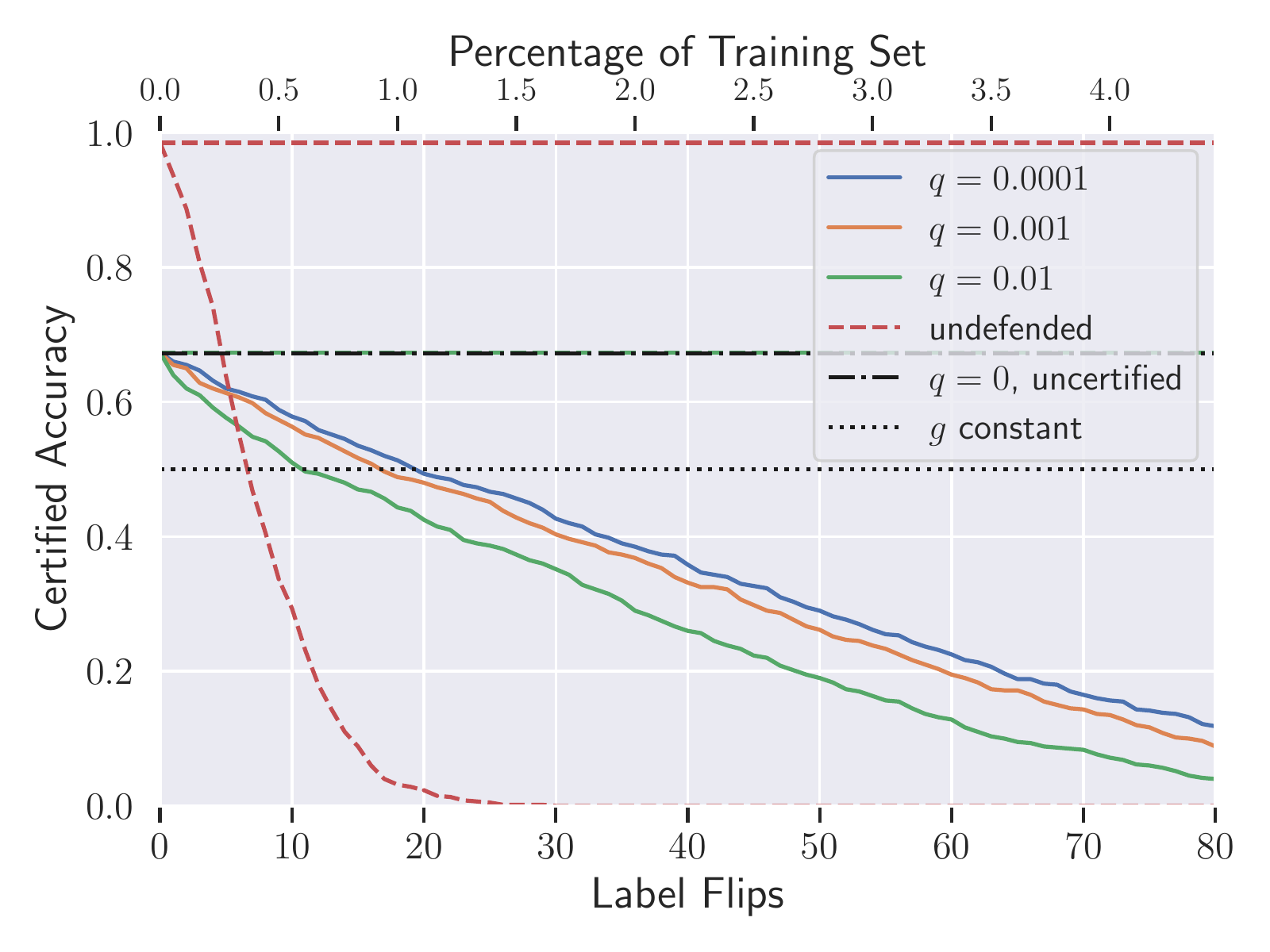}
    \caption{Dogfish test set certified accuracy using features learned with RICA \citep{UnsupervisedFeatures}. While not as performant as the pre-trained features, our classifier still achieves reasonable certified accuracy---note that the certified lines are lower bounds, while the undefended line is an upper bound. As demonstrated in the main body, deep unsupervised features significantly boost performance, but require a larger dataset.}
    \label{fig:dogfish-rand}
\end{figure}

\begin{figure} [h]
    \centering
    \includegraphics[width=.95\linewidth]{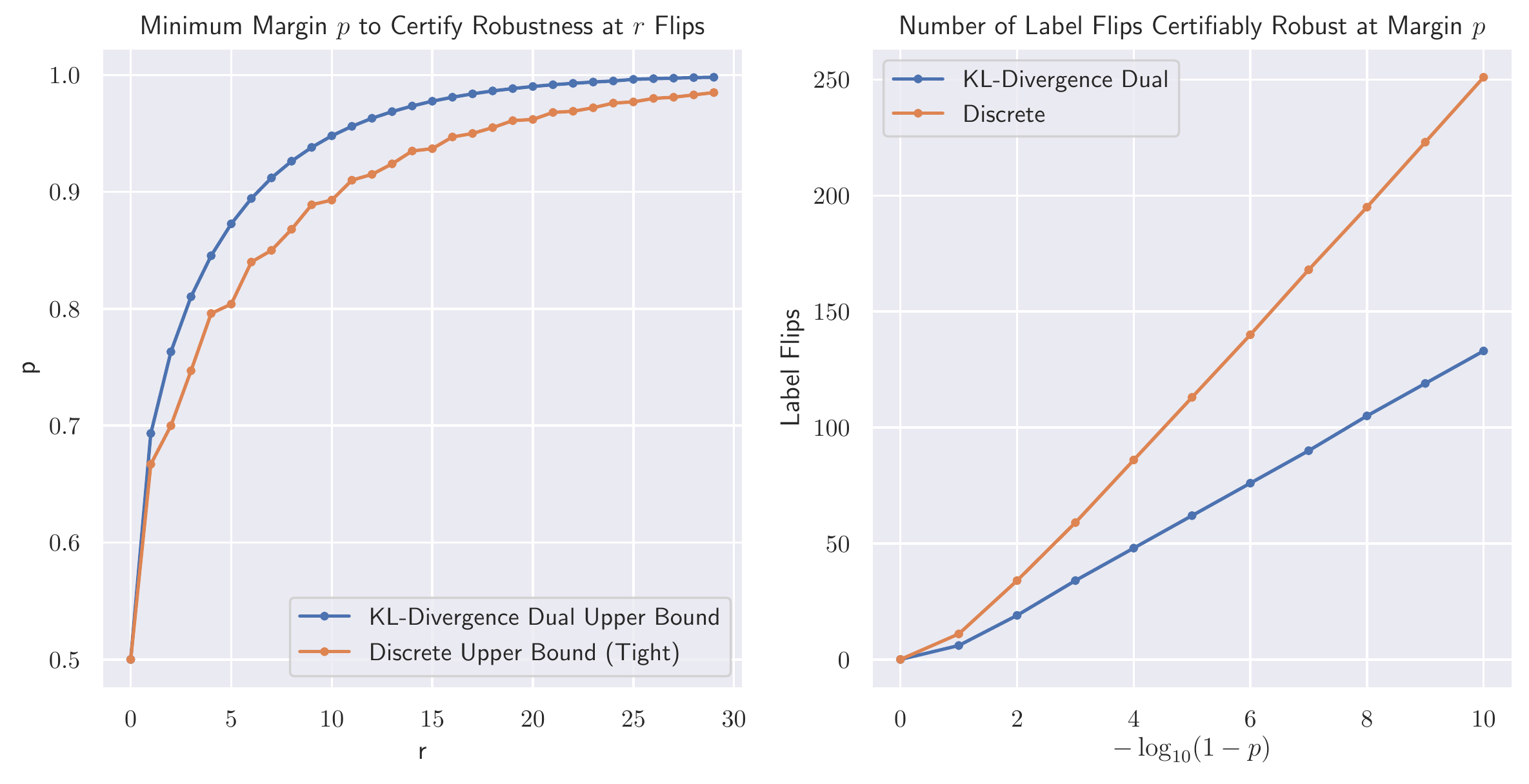}
    \caption{\textbf{Left:} Required margin $p$ to certify a given number of label flips using the generic KL bound (\ref{eq:kl-bound}) versus the tight discrete bound (\ref{eq:stratified-bound}). \textbf{Right:} The same comparison, but inverted, showing the certifiable robustness for a given margin. The tight bound certifies robustness to approximately twice as many label flips. Both plots are with $q=0.4$.}
    \label{fig:duality-gap}
\end{figure}

\end{document}